\documentclass[preprint,12pt,authoryear]{elsarticle}
\pdfoutput=1
\usepackage[T1]{fontenc}
\usepackage[utf8]{inputenc}
\usepackage{lmodern}
\usepackage{amsmath,amssymb,amsthm}
\usepackage{graphicx}
\usepackage{longtable,booktabs,array,calc}
\usepackage{xcolor}
\usepackage[a4paper,margin=2.5cm]{geometry}
\usepackage{caption}
\usepackage{float}
\usepackage{etoolbox}
\AtBeginEnvironment{longtable}{\footnotesize\setlength{\tabcolsep}{4.5pt}}
\makeatletter
\patchcmd\longtable{\par}{\if@noskipsec\mbox{}\fi\par}{}{}
\def\maxwidth{\ifdim\Gin@nat@width>\linewidth\linewidth\else\Gin@nat@width\fi}
\def\maxheight{\ifdim\Gin@nat@height>\textheight\textheight\else\Gin@nat@height\fi}
\makeatother
\setkeys{Gin}{width=\maxwidth,height=\maxheight,keepaspectratio}
\providecommand{\tightlist}{\setlength{\itemsep}{0pt}\setlength{\parskip}{0pt}}

\usepackage[hidelinks]{hyperref}
\usepackage{xurl}
\journal{Medical Image Analysis}
\begin{document}
\begin{frontmatter}
\title{Quasi-Binarized Autoencoders: An Architecture-Independent Information Bottleneck for Medical Image Anomaly Detection}
\author[a]{Shouhei Hanaoka\corref{cor1}}
\ead{hanaokalog@gmail.com}
\cortext[cor1]{Corresponding author.}
\author[b]{Takahiro Nakao}
\author[b,c]{Atsushi Takamatsu}
\author[b]{Takeharu Yoshikawa}
\author[a]{Osamu Abe}
\affiliation[a]{organization={Department of Radiology, The University of Tokyo Hospital},addressline={7-3-1 Hongo, Bunkyo-ku},city={Tokyo},postcode={113-8655},country={Japan}}
\affiliation[b]{organization={Department of Computational Diagnostic Radiology and Preventive Medicine, The University of Tokyo Hospital},addressline={7-3-1 Hongo, Bunkyo-ku},city={Tokyo},postcode={113-8655},country={Japan}}
\affiliation[c]{organization={Department of Radiology, Kanazawa University Hospital},addressline={13-1 Takara-machi},city={Kanazawa},state={Ishikawa},postcode={920-8641},country={Japan}}
\begin{abstract}
Unsupervised anomaly detection, which learns only from normal images, is a central task in medical image analysis and remains an open problem. Reconstruction-based methods pass an image through an encoder--decoder network trained on normal data and detect anomalies from the residual between the image and its reconstruction. This works only if the information passed from the encoder to the decoder is limited; otherwise the network learns an identity mapping and reconstructs anomalies too. This limit is usually imposed through architectural choices, tuned per dataset, that cannot be stated in bits. We introduce the quasi-binarizing (QB) layer, which squashes each latent element into \([0,1]\) and adds Laplace noise of scale \(1/\varepsilon\). Each element is then \(\varepsilon\)-locally differentially private, and the mutual information between an image and its reconstruction is bounded by a quantity that depends only on \(\varepsilon\) and the number of QB elements, whatever the encoder and decoder. Placing a QB layer on every encoder--decoder path, including all skip connections, we build QBAE, a seven-level attention U-Net with 32,768 QB elements. On the seven datasets of the MedIAnomaly benchmark, QBAE with one architecture and one configuration reaches a mean image-level AUROC of 0.828, the highest among methods that do not adapt to each dataset, and the best reported results on BraTS2021 (AUROC 0.911, pixel-level AP 0.838). The noise is kept at test time, so that every reconstruction satisfies the bound. Without input corruption, the bottleneck alone prevents identity collapse (mean AUROC 0.805 vs.~0.590). Code is available at https://github.com/hanaokalog/MedIAnomalyQB.
\end{abstract}
\end{frontmatter}

\hypertarget{introduction}{%
\section{Introduction}\label{introduction}}

Unsupervised anomaly detection aims to flag images that deviate from a model of normal anatomy, using only normal training data. This setting fits medical imaging well. Normal studies are plentiful, abnormal ones are diverse and costly to annotate, and the full range of pathologies can never be listed in advance. Many paradigms have been proposed, including image reconstruction, self-supervised pretext tasks, and feature embedding. Yet the recent MedIAnomaly benchmark (Cai et al., 2025) showed that reconstruction-based autoencoders (AEs) remain among the most competitive and versatile approaches across seven datasets and five imaging modalities. They are especially strong for pixel-level localization, where feature-embedding methods built on natural-image backbones struggle.

An AE detects anomalies only if it reconstructs normal images well and abnormal images poorly. This works only when the information that passes from the encoder to the decoder is limited. Otherwise the network learns a near-identity mapping and reconstructs lesions as faithfully as normal tissue (Cai et al., 2024). In practice this limit is imposed through the \emph{architecture}. Designers shrink the latent dimension, remove skip connections, or reduce depth, and then tune these choices per dataset. This practice has two drawbacks.

\begin{itemize}
\tightlist
\item
  \textbf{Expressiveness and information limit are coupled.} Components that make reconstructions sharper, such as U-Net skip connections (Ronneberger et al., 2015), attention, and modern up/down-sampling blocks (Chen et al., 2025), also widen the path through which identity can leak. Anomaly-detection AEs therefore stay deliberately small and shallow, while the rest of deep learning has moved on.
\item
  \textbf{The limit is not measured.} A real-valued latent vector of any dimension can in principle carry unbounded information. The latent dimension is only an indirect proxy, and the amount of information an AE actually transmits can neither be stated in bits nor compared across architectures.
\end{itemize}

Existing remedies address these issues only in part.

\begin{itemize}
\tightlist
\item
  Variational AEs (Kingma and Welling, 2014) quantify information through a KL term, but they tie the bottleneck to a prior and tend to produce blurry reconstructions.
\item
  Memory-augmented (Gong et al., 2019) and vector-quantized (van den Oord et al., 2017) AEs restrict the latent to a learned codebook. However, the codebook size is again an architectural choice, and skip connections bypass it.
\item
  Denoising AEs (Vincent et al., 2008; Kascenas et al., 2022) take a complementary route. They corrupt the \emph{input} so that the network must learn to map corrupted images back onto the normal manifold. This yields the strongest image-level results on brain MRI in MedIAnomaly. Yet input corruption shapes \emph{what} the network learns, not \emph{how much} information it can transmit.
\end{itemize}

\textbf{We propose to separate the information bottleneck from the architecture.} We introduce the \emph{quasi-binarizing (QB) layer}. A QB layer squashes each element of the latent tensor into {[}0, 1{]} with a sigmoid and adds Laplace noise of scale 1/\ensuremath{\varepsilon}. To remain readable despite this noise, the outputs are driven toward 0 or 1 during training, so that the layer acts as a soft, noisy binarizer; hence its name (Section 3.2.1). The construction comes from differential privacy (DP), a framework for releasing the output of a computation while provably limiting what that output reveals about its input (Dwork et al., 2006). In DP terms, each element has \emph{sensitivity} one, i.e.~its value can change by at most one whatever the input (a property of the function, unrelated to the sensitivity of a diagnostic test), and adding Laplace noise of scale 1/\ensuremath{\varepsilon} is exactly the Laplace mechanism. Every element is therefore \ensuremath{\varepsilon}-locally differentially private (Duchi et al., 2013). This yields two bounds on the information an element carries about the input image.

\begin{itemize}
\tightlist
\item
  \textbf{Privacy bound.} The mutual information is at most min\{\ensuremath{\varepsilon}, \ensuremath{\varepsilon}\textsuperscript{2}\} nats (Cuff and Yu, 2016).
\item
  \textbf{Amplitude bound.} Because each element is confined to {[}0, 1{]}, the mutual information is also at most \ensuremath{\tfrac{1}{2}} ln(2\ensuremath{\pi}e(\ensuremath{\tfrac{1}{4}} + 2/\ensuremath{\varepsilon}\textsuperscript{2})) \ensuremath{-} ln(2e/\ensuremath{\varepsilon}) nats (Cover and Thomas, 2006). This bound grows only logarithmically in \ensuremath{\varepsilon}, for example to about 2 bits per element at \ensuremath{\varepsilon} = 10.
\end{itemize}

The noise is drawn independently for each element. The information the decoder receives is therefore bounded by the sum of these per-element bounds, whatever the depth, width, or connectivity of the encoder and decoder. We place a QB layer on \emph{every} path from the encoder to the decoder, including all skip connections. The resulting budget depends only on \ensuremath{\varepsilon} and the number of QB elements. We keep the noise at test time, so that every reconstruction used for detection is obtained under this budget, and the bottleneck can be stated, and audited, in bits.

Once the information limit no longer depends on the architecture, the encoder and decoder can be made as expressive as reconstruction quality requires. Our Quasi-Binarized Autoencoder (QBAE) is a seven-level U-Net with full-resolution skip connections. It uses DC-AE residual autoencoding blocks (Chen et al., 2025), cross-attention gates that mix each skip with the upsampled path, and attention blocks in the decoder. Every skip connection passes through a QB layer, with the number of QB elements halved at each level, and a fully connected layer at the 2 \ensuremath{\times} 2 bottom feeds a final QB layer. In total, 32,768 QB elements feed the decoder. Classical design rules would call this latent far too large for anomaly detection. Under the QB bottleneck it is not: at \ensuremath{\varepsilon} = 10, these elements carry at most 6.5 \ensuremath{\times} 10\textsuperscript{4} bits, half the 131,072 bits of a raw 8-bit 128 \ensuremath{\times} 128 input image.

Because the budget is set by \ensuremath{\varepsilon} alone, one fixed network can be adapted to datasets with very different anomalies by changing a single scalar with a direct meaning, the number of bits allowed through, instead of redesigning the latent space or the skip connections.

Our contributions are:

\begin{enumerate}
\def\labelenumi{\arabic{enumi}.}
\tightlist
\item
  \textbf{The QB layer.} It is a bounded-input Laplace noise layer that gives every encoder--decoder path an information budget controlled by \ensuremath{\varepsilon}, with an \ensuremath{\varepsilon}-LDP guarantee and closed-form bounds on the mutual information per element.
\item
  \textbf{QBAE.} It is an anomaly-detection U-Net in which every encoder--decoder path, including all skip connections, passes through a QB layer. This allows modern high-capacity components without identity leakage.
\item
  \textbf{A systematic evaluation} on all seven MedIAnomaly datasets at 128 \ensuremath{\times} 128 resolution, with one network and a sweep of \ensuremath{\varepsilon} over almost two orders of magnitude. With a single configuration for all datasets, QBAE reaches the highest mean AUROC (0.828) among methods that do not adapt to each dataset, and an image-level AUROC of 0.911 and a pixel-level AP of 0.838 on BraTS2021, exceeding the best reported results (0.859 and 0.755). We show that the best \ensuremath{\varepsilon} differs systematically between datasets, that performance is insensitive to \ensuremath{\varepsilon} near its optimum, and that without input corruption the QB bottleneck alone prevents the identity collapse of the network.
\item
  \textbf{Open-source code.} The implementation, training scripts and configurations are available at \url{https://github.com/hanaokalog/MedIAnomalyQB}.
\end{enumerate}

\hypertarget{related-work}{%
\section{Related Work}\label{related-work}}

\textbf{Reconstruction-based anomaly detection in medical imaging.}
AEs and their variants are the most widely used family for unsupervised anomaly detection in medical images (Baur et al., 2021). Examples include adversarial models such as f-AnoGAN (Schlegl et al., 2019), context-encoding VAEs (Zimmerer et al., 2018), AEs with uncertainty estimation (Mao et al., 2020), and perceptual AEs (Shvetsova et al., 2021). Denoising AEs (Kascenas et al., 2022) corrupt brain MR images with coarse, spatially correlated noise, and this simple scheme rivals far more complex models. More recently, diffusion models have been used to ``heal'' anomalies (Wolleb et al., 2022; Wyatt et al., 2022; Pinaya et al., 2022), at a much higher inference cost.

Two benchmarks now enable a fair comparison across modalities. BMAD (Bao et al., 2024) covers six datasets, and MedIAnomaly (Cai et al., 2025) evaluates more than 30 methods on seven datasets under a unified protocol. MedIAnomaly found that well-tuned AEs remain highly competitive. Its companion analysis (Cai et al., 2024) argued, from an information-theoretic viewpoint, that the latent size, and hence the information passed to the decoder, is the key factor in AE-based detection.

Training-free methods that score densities of pretrained features have also been benchmarked on MedIAnomaly. MSDE (Kar et al., 2026) refines embeddings from a pretrained network (ResNet-18 or a self-supervised AnatPaste encoder, chosen per dataset) by mean shift before Gaussian density scoring. It reaches high image-level AUROC on chest X-ray and brain-tumor MRI, but it falls well behind reconstruction-based methods on BraTS2021 (0.736 vs.~0.859 for DAE).

In industrial anomaly detection, the failure mode in which an over-capable reconstruction network copies anomalies to its output is known as the ``identical shortcut'' (You et al., 2022). The reconstruction-based works above all control the information that reaches the decoder indirectly, through architectural choices such as latent size, depth, or the absence of skip connections. QBAE instead controls it with an explicit, architecture-independent budget.

\textbf{Limiting and measuring information in autoencoders.}
The information bottleneck (Tishby et al., 1999) and its variational form (Alemi et al., 2017) formalize the trade-off between compressing the input and keeping what is relevant. In VAEs (Kingma and Welling, 2014) and \ensuremath{\beta}-VAEs (Higgins et al., 2017), the KL term is an upper bound on the rate. However, this rate is measured relative to a learned prior, and skip connections fall outside it (Alemi et al., 2018). Discrete bottlenecks such as vector quantization (van den Oord et al., 2017) and memory modules (Gong et al., 2019) limit the latent to a codebook, so their capacity follows from the codebook size, which is again an architectural choice. Sparse AEs penalize mean activations with a KL term (Ng, 2011), but they bound no information-theoretic quantity.

The idea closest to the QB layer comes from learned hashing and compression. Semantic hashing (Salakhutdinov and Hinton, 2009; Kaiser and Bengio, 2018) adds noise \emph{before} a sigmoid to push the codes toward binary values. Learned image compression (Ballé et al., 2017) adds uniform noise as a differentiable proxy for quantization and measures the rate in bits. Neural joint source--channel coding (Choi et al., 2019) trains codes to survive a noisy binary channel. The QB layer differs from all of these in three ways:

\begin{itemize}
\tightlist
\item
  It adds noise \emph{after} a bounded squashing. This is what yields a formal differential-privacy guarantee and closed-form bounds on the information per element. Because the noise has the same scale everywhere, values are easiest to tell apart when they lie at the two ends of \([0,1]\); training therefore drives the outputs toward 0 or 1, and the output of the QB layer becomes nearly binary.
\item
  It is placed on every encoder--decoder path, including the skips.
\item
  Its purpose is to \emph{withhold} information, to prevent identity mapping, rather than to transmit it efficiently.
\end{itemize}

\textbf{Differential privacy and information.}
Differential privacy (DP) is a formal framework for releasing the output of a computation, called a \emph{mechanism}, while limiting how much that output can reveal about any single input record: a mechanism is \ensuremath{\varepsilon}-DP if changing one record changes the probability of any output by at most a factor \(e^{\varepsilon}\) (Dwork et al., 2006). Smaller \ensuremath{\varepsilon} thus means less can be inferred about the input from the output. The \emph{sensitivity} of a function is the largest change in its value that a change of one record can cause. The Laplace mechanism (Dwork et al., 2006) makes a function with bounded sensitivity \ensuremath{\varepsilon}-differentially private by adding Laplace noise scaled to that sensitivity. In the local model, each record is privatized before it leaves the data owner (Duchi et al., 2013). \ensuremath{\varepsilon}-DP implies a bound on mutual information (McGregor et al., 2010; Cuff and Yu, 2016). In deep learning, DP is mostly applied to \emph{training data} through gradient perturbation (Abadi et al., 2016). Noise injected into intermediate activations has been studied to protect inference-time privacy (Mireshghallah et al., 2020). To our knowledge, the Laplace mechanism has not been used as an information bottleneck for anomaly detection. In QBAE, the privacy parameter \ensuremath{\varepsilon} directly sets how much of each image the decoder may ``see''.

\textbf{Architectural components and pretrained features.}
QBAE builds on the U-Net (Ronneberger et al., 2015) with attention gating of skip connections (Oktay et al., 2018). Its gates follow CBAM (Woo et al., 2018), extended to cross-attention between the skip and the upsampled path. The encoder and decoder use the residual autoencoding blocks of DC-AE (Chen et al., 2025), which pair space-to-channel shortcuts without parameters with learned residuals. The decoder and the bottom of the network use channel attention in the style of Restormer (Zamir et al., 2022) and spatial self-attention. None of these components is specific to anomaly detection. The point of QBAE is that, because every path to the decoder passes through a QB layer, they can be adopted without re-introducing an identity shortcut.

Like perceptual AEs (Shvetsova et al., 2021) and the AE-Perceptual baseline of MedIAnomaly, QBAE trains with, and scores by, a relative perceptual loss computed on ImageNet-pretrained VGG19 features (Simonyan and Zisserman, 2015; Johnson et al., 2016). It therefore also relies on external pretrained weights. In contrast to feature-density methods such as MSDE, however, these features enter only the loss and the anomaly score. The representation that crosses the bottleneck is learned from normal medical images alone.

\hypertarget{method}{%
\section{Method}\label{method}}

\hypertarget{problem-setting-and-overview}{%
\subsection{Problem setting and overview}\label{problem-setting-and-overview}}

We consider the standard unsupervised setting. A training set \(\mathcal{D}=\{x^{(m)}\}_{m=1}^{M}\) contains only normal images \(x\in\mathbb{R}^{C\times H\times W}\). At test time, the model must produce a pixel-wise anomaly map \(A(x)\in\mathbb{R}^{H\times W}\) and an image-level score \(s(x)\in\mathbb{R}\) for images that may contain unseen abnormalities. Like other reconstruction-based methods, QBAE trains a network \(F\) to reconstruct normal images and derives \(A\) and \(s\) from the discrepancy between \(x\) and \(\hat{x}=F(x)\).

QBAE (Fig. 1) is a seven-level U-Net-type autoencoder in which \textbf{every} tensor passed from the encoder to the decoder first goes through a quasi-binarizing (QB) layer. At each encoder level \(l\in\{1,\dots,6\}\) (resolutions \(128^2\) down to \(4^2\)), the skip connection is projected to \(k_l=2^{\,l-1}\) channels per pixel and passed through a QB layer, so that the number of QB elements halves from level to level: \(16{,}384, 8{,}192, \dots, 512\). At the deepest level, the \(2\times2\) feature map is flattened and mapped by a fully connected layer to a further \(512\) QB elements. This gives

\[
N=\sum_{l=1}^{6}2^{\,l-1}\,\frac{H W}{4^{\,l-1}} + 512
 = 16{,}384+8{,}192+4{,}096+2{,}048+1{,}024+512+512 = 32{,}768 = 2^{15}
\]

QB elements in total, for \(H=W=128\). The same network, with the same \(N\), is used for all datasets in this work; only \(\varepsilon\) is changed. Let \(h\in\mathbb{R}^{N}\) denote the concatenated pre-activations produced by the encoder \(E\). Let \(\tilde z\in\mathbb{R}^{N}\) denote the corresponding QB outputs. The decoder \(D\) produces \(\hat x = D(\tilde z)\). The encoder and decoder are otherwise unconstrained. Section 3.2 shows that the QB layers alone bound how much information about \(x\) can reach \(\hat x\). Section 3.3 describes the architecture and shows that no path from \(x\) to \(\hat x\) avoids a QB layer. During training, the encoder receives a corrupted image \(\mathcal{C}(x)\), and the network is trained to reconstruct the clean \(x\) (Section 3.4). Section 3.5 defines the anomaly scores.

\begin{figure}
\centering
\includegraphics[width=0.92\textwidth,height=\textheight]{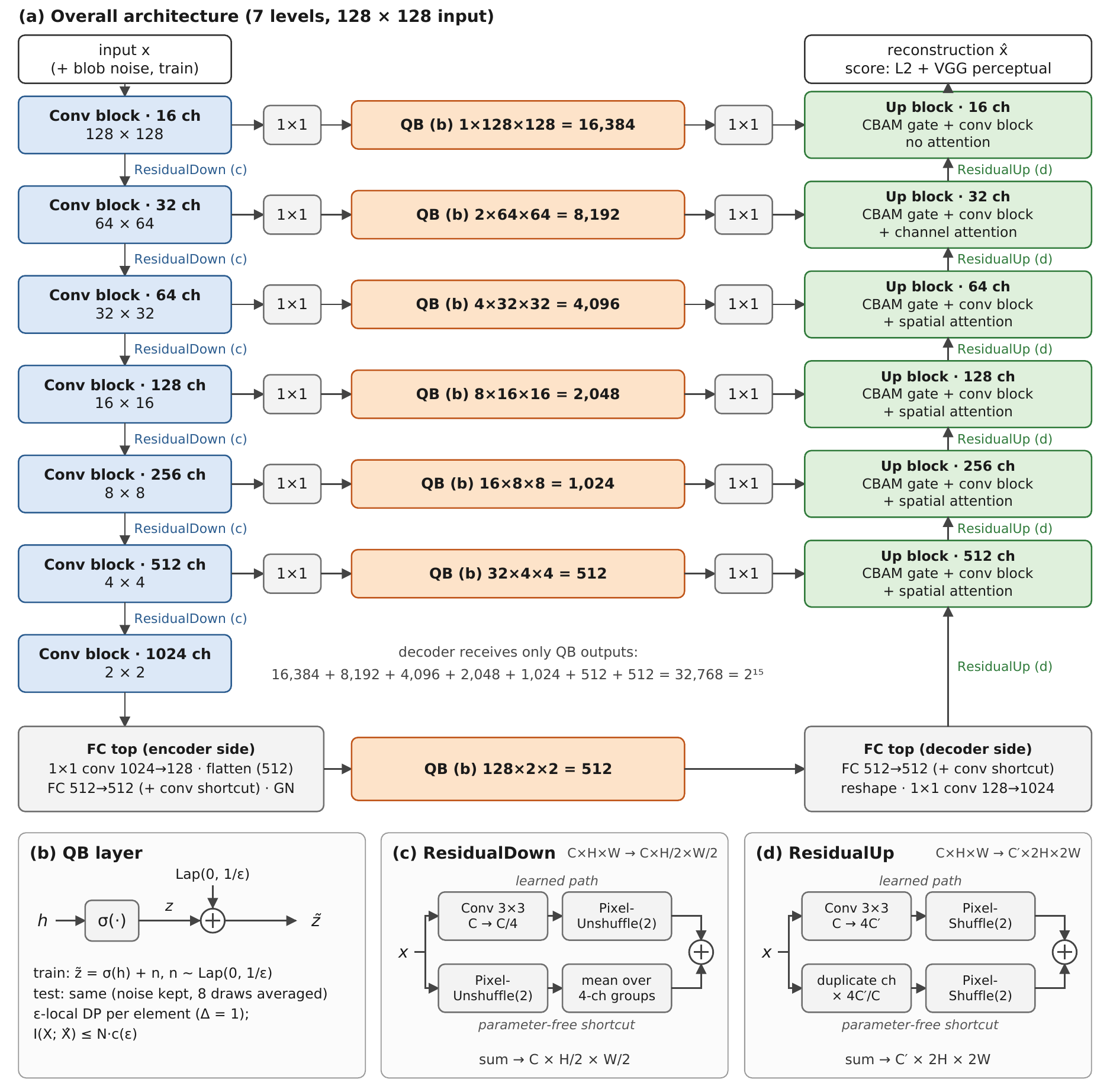}
\caption{\textbf{Fig. 1.} Overview of QBAE. (a) A seven-level U-Net in which each of the six skip connections and the fully connected bottom passes through a QB layer; the decoder receives only QB outputs (32,768 elements). (b) QB layer. (c), (d) DC-AE residual down- and upsampling. Details are given in Appendix B.}
\end{figure}

\hypertarget{the-quasi-binarizing-layer}{%
\subsection{The quasi-binarizing layer}\label{the-quasi-binarizing-layer}}

\hypertarget{definition}{%
\subsubsection{Definition}\label{definition}}

A QB layer with privacy parameter \(\varepsilon>0\) acts element-wise on a pre-activation vector \(h\in\mathbb{R}^N\). It first squashes each element into the unit interval,

\[
z_i=\sigma(h_i)\in[0,1],
\]

and, during training, adds independent Laplace noise,

\[
\tilde z_i = z_i + n_i,\qquad n_i \overset{\text{i.i.d.}}{\sim}\operatorname{Laplace}(0,\,b),\quad b = 1/\varepsilon .
\tag{1}
\]

The noise is resampled for every sample, every element, and every forward pass, independently of the input. Since the noise is additive, gradients pass through the layer unchanged, and no straight-through estimator is required.

The noise is kept at test time, so that the network used for detection is the same noisy mechanism as in training (Section 3.5).

Two limits help with interpretation. As \(\varepsilon\to\infty\) the noise vanishes, and the QB layer reduces to a sigmoid. In our experiments this case is approximated by \(\varepsilon=10^8\) and serves as the \emph{no-bottleneck} control. As \(\varepsilon\to0\) the noise swamps the signal, and no information is transmitted.

The name reflects the effect of the noise. To survive noise of scale \(b\), the encoder is driven to place each \(z_i\) near the ends of \([0,1]\), so that its value can still be read after noise is added. The learned code thus becomes \emph{quasi}-binary (Appendix D).

\hypertarget{information-budget}{%
\subsubsection{Information budget}\label{information-budget}}

\textbf{In short: the amount of information that can pass through QBAE has a rigorous upper bound, which depends only on \ensuremath{\varepsilon} and on the number of QB elements and not on the architecture of the encoder or the decoder.}

The intuition is simple. Whatever the encoder computes, each QB element can only send a value between 0 and 1, and during training this value reaches the decoder blurred by noise of scale \(1/\varepsilon\). A receiver can therefore tell apart only a limited number of levels, roughly of the order of \(\varepsilon\), no matter how cleverly the encoder chooses them. Each element thus behaves like a noisy dial with a fixed resolution. The decoder sees the image only through \(N\) such dials, so its view of the image is limited by the number of dials and their resolution, not by how deep or wide the network around them is. This is the same mechanism that differential privacy uses to limit what a released statistic reveals about an individual (the Laplace mechanism; Dwork et al., 2006). Here, the ``individual'' is the input image.

This intuition can be made exact. We state the result informally here; the formal statements and proofs are given in Appendix A.

\textbf{Theorem 1 (information budget, informal).}
Assume that every path from the input to the output passes through a QB layer (Section 3.3 shows that QBAE satisfies this). Then, during training and at test time, the mutual information between the input image \(X\) and the reconstruction \(\hat X\) satisfies
\[
I(X;\hat X)\;\le\;N\cdot c(\varepsilon),
\]
where \(c(\varepsilon)\) is an explicit per-element bound that depends on \(\varepsilon\) only (Appendix A, Eq. (A.4)).

The bound holds for any encoder and decoder, for normal and abnormal images alike, and with or without input corruption. For large \(\varepsilon\), \(c(\varepsilon)\) grows only logarithmically, roughly as \(\log_2\varepsilon\) bits. Table 1 gives the budgets for the settings used in this work (\(\varepsilon\) on a \(\sqrt{10}\) grid). To put them in perspective, a raw 8-bit \(128\times128\) single-channel image contains \(131{,}072\) bits.

\textbf{Table 1.} Upper bounds on the information transmitted through QBAE (\(N=32{,}768\) QB elements).

\begin{longtable}[]{@{}
  l
  l
  l
  l@{}}
\toprule\noalign{}
Setting & Per element & Total budget & Relative to a raw 8-bit \(128^2\) image \\
\midrule\noalign{}
\endhead
\bottomrule\noalign{}
\endlastfoot
\(\varepsilon=0.3\) & 0.11 bits & \(3.7\times10^{3}\) bits & 0.03 \\
\(\varepsilon=1\) & 0.19 bits & \(6.2\times10^{3}\) bits & 0.05 \\
\(\varepsilon=3\) & 0.65 bits & \(2.1\times10^{4}\) bits & 0.16 \\
\(\varepsilon=10\) & 1.98 bits & \(6.5\times10^{4}\) bits & 0.50 \\
\(\varepsilon=30\) & 3.52 bits & \(1.15\times10^{5}\) bits & 0.88 \\
\(\varepsilon=100\) & 5.25 bits & \(1.72\times10^{5}\) bits & 1.31 \\
\end{longtable}

Because the budget is set by a single scalar, it can be adapted to a dataset without changing the network: across the range of Table 1 it spans almost two orders of magnitude.

Theorem 1 applies to the \emph{noisy} QB layers. Removing the noise at test time would make the layers noiseless and their capacity unbounded in principle. We therefore keep the noise at test time and average the anomaly scores over several noise draws (Section 3.5), so that every reported result is obtained under the stated budget.

\hypertarget{practical-remarks}{%
\subsubsection{Practical remarks}\label{practical-remarks}}

\textbf{Saturation.} A sigmoid that saturates early would make the layer hard to train. At the bottom QB layer, we therefore apply a learnable parametrized arctangent, \(\mathrm{PA}(h)=\alpha\arctan(\beta h+\gamma)+\delta\), before \(\sigma\). Here \(\alpha,\beta,\gamma,\delta\) are learnable scalars, and \(\beta\) is initialized to \(0.01\), so that \(\sigma(\mathrm{PA}(h))\) starts close to \(\tfrac12\). The QB layers of the skip connections take the output of a \(1\times1\) convolution directly.

\textbf{Sparsity.} Optionally, a sparsity penalty can push the mean activation of each element toward a target \(\rho\) (Section 3.4). The bounds above do not rely on this penalty.

\textbf{Normalization.} All normalization layers in QBAE operate within each sample (GroupNorm). Batch statistics therefore cannot carry information from one sample to another around the QB layers.

\hypertarget{network-architecture}{%
\subsection{Network architecture}\label{network-architecture}}

The architecture of QBAE is not where its bottleneck comes from, and we therefore keep its description short. The network is chosen to reconstruct normal images well; how much information it may pass is fixed separately by the QB layers (Section 3.2). All experiments use the same network, shown in Fig. 1. Layer-level details are given in Appendix B.

\textbf{Encoder and decoder.} QBAE is a seven-level U-Net operating on \(128\times128\) inputs, with feature widths \(16, 32, \dots, 1024\) at resolutions \(128^2\) down to \(2^2\). Each level uses a residual convolution block (two \(3\times3\) convolutions with Swish activations (Ramachandran et al., 2017) and GroupNorm (Wu and He, 2018)). Down- and upsampling follow the residual autoencoding of DC-AE (Chen et al., 2025), in which a parameter-free space-to-channel rearrangement is added to a learned convolution. In the decoder, each skip connection is merged with the upsampled path by a cross-attention gate (Woo et al., 2018), and the decoder levels at \(64^2\) and below contain an attention block (Zamir et al., 2022).

\textbf{QB layers.} Every tensor that the decoder receives from the encoder passes through a QB layer. At each of the six skip connections, a \(1\times1\) convolution projects the encoder features to \(2^{\,l-1}\) channels per pixel at level \(l\), a QB layer is applied, and a second \(1\times1\) convolution maps the result back to the decoder width. Because the number of pixels falls by a factor of four per level while the number of channels per pixel doubles, the number of QB elements halves from one level to the next (Table 2). At the bottom, the \(2\times2\) feature map is flattened and passed through a fully connected layer to a final QB layer of 512 elements, the same number as the deepest skip connection. The fully connected layer gives the bottom path position-specific weights, which the convolutional levels lack. In total, the decoder receives \(N=32{,}768\) QB elements.

\textbf{Table 2.} QB elements per level (\(128\times128\) input).

\begin{longtable}[]{@{}lllll@{}}
\toprule\noalign{}
Level & Resolution & Channels per pixel & QB elements & Share of \(N\) \\
\midrule\noalign{}
\endhead
\bottomrule\noalign{}
\endlastfoot
1 & \(128^2\) & 1 & 16,384 & 50.0\% \\
2 & \(64^2\) & 2 & 8,192 & 25.0\% \\
3 & \(32^2\) & 4 & 4,096 & 12.5\% \\
4 & \(16^2\) & 8 & 2,048 & 6.3\% \\
5 & \(8^2\) & 16 & 1,024 & 3.1\% \\
6 & \(4^2\) & 32 & 512 & 1.6\% \\
bottom & \(2^2\) (fully connected) & -- & 512 & 1.6\% \\
total & & & 32,768 & 100\% \\
\end{longtable}

\textbf{No path avoids a QB layer.} Theorem 1 requires that the reconstruction depends on the input only through the QB outputs. QBAE satisfies this by construction. The residual shortcuts of the convolution blocks and of the DC-AE resampling layers stay inside the encoder or inside the decoder. The cross-attention gates combine only QB outputs with decoder features. All normalization layers act within a single sample, so no information can pass from one image to another through batch statistics. Hence every path from \(x\) to \(\hat x\) in the computation graph contains a QB layer, and the bound of Section 3.2.2 applies to the network as a whole, whatever the capacity of the encoder and the decoder.

\hypertarget{training}{%
\subsection{Training}\label{training}}

\textbf{Input corruption.} During training, the encoder receives a corrupted image \(\mathcal{C}(x)=x+\eta\) and the network is trained to reconstruct the clean \(x\), as in the denoising autoencoder (DAE) of Kascenas et al.~(2022). Our corruption is a modified version of their coarse noise (Kascenas et al., 2022, 2023), which is low-resolution Gaussian noise upsampled to the image size. Instead of covering the whole image, our \(\eta\) consists of smooth, localized blobs of random size. It is the product of two Gaussian-smoothed white-noise fields with random smoothing radii: one field sets the texture, and the other is thresholded and raised to a random power so that only a few patches remain non-zero. The product is scaled by the standard deviation of the image. The exact generator is given in Appendix C. Input corruption and the QB layers play different roles: the corruption shapes \emph{what} the network learns, namely to map locally altered images back to normal ones, whereas the QB layers limit \emph{how much} information about the input can reach the decoder. Section 5 shows that the two are complementary.

\textbf{Loss.} The network is trained with
\[
\mathcal{L}=\frac{1}{HW}\,\lVert x-\hat x\rVert_2^2+\lambda_p\,\mathcal{L}_{\mathrm{perc}}(x,\hat x)+\lambda_{\mathrm{KL}}\sum_{i=1}^{N}\mathrm{KL}\!\left(\rho\,\Vert\,\bar z_i\right),
\tag{2}
\]
where \(\hat x\) is the reconstruction of \(\mathcal{C}(x)\). The perceptual term \(\mathcal{L}_{\mathrm{perc}}\) is the relative-perceptual-\(L_1\) loss used by the perceptual autoencoder (AE-PL) of Shvetsova et al.~(2021): an \(L_1\) distance between VGG-19 (Simonyan and Zisserman, 2015) relu4\_2 features with ImageNet (Deng et al., 2009) weights of \(x\) and \(\hat x\), with the features standardized per channel and the distance divided by the mean feature magnitude of \(x\). The only change is that, during training, \(x\) and \(\hat x\) are jointly shifted by up to 8 pixels before the feature extraction. The last term is an optional sparsity penalty (Ng, 2011): \(\bar z_i\) is the mean of \(z_i=\sigma(h_i)\) over the mini-batch, and \(\mathrm{KL}(\rho\Vert\bar z_i)\) is the Kullback--Leibler divergence between Bernoulli distributions with means \(\rho\) and \(\bar z_i\). It encourages each element to be active for only a fraction \(\rho\) of the images, so that the code becomes sparse as well as quasi-binary. We report results with and without this term. None of the bounds in Section 3.2 depends on it.

\hypertarget{anomaly-scores}{%
\subsection{Anomaly scores}\label{anomaly-scores}}

At test time, the corruption is switched off and the QB layers keep their noise (Section 3.2.1). The pixel-wise anomaly map combines the squared reconstruction error with the perceptual error map,
\[
A(x)=(x-\hat x)^2+\lambda_p\,\mathcal{U}\big(D_{\mathrm{perc}}(x,\hat x)\big),
\tag{3}
\]
where \(D_{\mathrm{perc}}\) is the channel-averaged relative \(L_1\) feature difference before spatial averaging and \(\mathcal{U}\) denotes bilinear upsampling to the image size. The second term is exactly the anomaly map of AE-PL (Shvetsova et al., 2021; Cai et al., 2025). The map \(A(x)\) is used for pixel-level evaluation. The image-level score \(s(x)\) is the spatial mean of this perceptual term alone, \(s(x)=\lambda_p\,\overline{\mathcal{U}(D_{\mathrm{perc}}(x,\hat x))}\), i.e.~the image-level score of AE-PL.

Because the noise is kept, the score depends on the noise draw. We therefore average \(A(x)\) and \(s(x)\) over \(K=8\) independent draws; \(K\) was fixed in advance and is the same for all datasets. Every test-time reconstruction is produced by exactly the noisy network analyzed in Theorem 1 and is thus obtained under the stated budget. Because the averaged score combines \(K\) independent passes, Proposition 3 bounds the information it carries about \(x\) by \(K\) times this budget.

\hypertarget{experimental-setup}{%
\section{Experimental setup}\label{experimental-setup}}

\hypertarget{datasets}{%
\subsection{Datasets}\label{datasets}}

We use all seven datasets of the MedIAnomaly benchmark (Cai et al., 2025) with its official training and test splits (Table 3). The training sets contain only normal images. The datasets cover five modalities and differ widely in the size and visibility of the abnormalities, from large tumors in brain MRI to subtle changes in retinal fundus photographs and histopathology patches. BraTS2021 is the only dataset with pixel-level annotations. All images are resized to \(128\times128\). We follow the benchmark protocol of downsampled inputs so that our results are comparable with the reported baselines, but use twice its default size of \(64\times64\).

\textbf{Table 3.} Datasets (MedIAnomaly splits).

\begin{longtable}[]{@{}
  l
  >{\raggedright\arraybackslash}p{3.6cm}
  l
  l
  l
  l
  l@{}}
\toprule\noalign{}
Dataset & Source & Modality & \begin{tabular}[b]{@{}l@{}}Train\\(normal)\end{tabular} & \begin{tabular}[b]{@{}l@{}}Test\\normal\end{tabular} & \begin{tabular}[b]{@{}l@{}}Test\\abnormal\end{tabular} & \begin{tabular}[b]{@{}l@{}}Pixel\\labels\end{tabular} \\
\midrule\noalign{}
\endhead
\bottomrule\noalign{}
\endlastfoot
RSNA & Shih et al.~(2019)\(^{a}\) & Chest X-ray & 3,851 & 1,000 & 1,000 & -- \\
VinDr-CXR & Nguyen et al.~(2022) & Chest X-ray & 4,000 & 1,000 & 1,000 & -- \\
Brain Tumor & Hamada (2025); Saleh et al.~(2020); Cheng et al.~(2015)\(^{b}\) & Brain MRI & 1,000 & 600 & 600 & -- \\
LAG & Li et al.~(2019) & Fundus photograph & 1,500 & 811 & 811 & -- \\
ISIC2018 & Codella et al.~(2019) & Dermoscopy & 6,705 & 909 & 603 & -- \\
Camelyon16 & Ehteshami Bejnordi et al.~(2017); Bao et al.~(2024)\(^{c}\) & Histopathology & 5,088 & 1,120 & 1,113 & -- \\
BraTS2021 & Baid et al.~(2021) & Brain MRI (FLAIR) & 4,211 & 828 & 1,948 & yes \\
\end{longtable}

\(^{a}\) RSNA Pneumonia Detection Challenge (2018), \url{https://www.rsna.org/education/ai-resources-and-training/ai-image-challenge/RSNA-Pneumonia-Detection-Challenge-2018}. \(^{b}\) Normal images from Br35H (Hamada, 2025) and Saleh et al.~(2020); glioma and meningioma images from Saleh et al.~(2020) and Cheng et al.~(2015). \(^{c}\) Pre-processed patches provided by Bao et al.~(2024).

\hypertarget{implementation-details}{%
\subsection{Implementation details}\label{implementation-details}}

QBAE is implemented in PyTorch on top of the MedIAnomaly codebase (Cai et al., 2025); our code is available at \url{https://github.com/hanaokalog/MedIAnomalyQB}. All experiments use the network of Section 3.3 without modification (\(N=32{,}768\) QB elements). Networks are trained with AdamW (Loshchilov and Hutter, 2019; learning rate \(10^{-3}\), weight decay \(10^{-3}\), batch size 128). The learning rate is increased linearly over the first 5 epochs and then decreased to \(10^{-5}\) with a cosine schedule (Loshchilov and Hutter, 2017); the global gradient norm is clipped at 1 (Pascanu et al., 2013). Training lasts 250 epochs (600 for Brain Tumor, following the benchmark). The perceptual weight is \(\lambda_p=0.1\). When the sparsity penalty is used, \(\rho=0.05\) and \(\lambda_{\mathrm{KL}}=10^{-6}\). The input corruption has unit strength relative to the image standard deviation. The VGG-19 network of the perceptual loss is evaluated in bfloat16 during training and in single precision at test time. No hyper-parameter other than \(\varepsilon\), the input corruption (on/off) and the sparsity penalty (on/off) is varied between experiments. Each run takes about 10--45 min on a single NVIDIA GH200 GPU, depending on the dataset.

We vary \(\varepsilon\) on a \(\sqrt{10}\) grid, \(\varepsilon\in\{0.3, 1, 3, 10, 30, 100\}\), which spans information budgets from \(3.7\times10^{3}\) to \(1.7\times10^{5}\) bits (Table 1). As a control without a bottleneck, we use \(\varepsilon=10^{8}\): the QB layers then reduce to a sigmoid, and the Laplace noise is negligible.

\hypertarget{evaluation-protocol}{%
\subsection{Evaluation protocol}\label{evaluation-protocol}}

We report the image-level AUROC as the main metric and the image-level AP as a secondary one. On BraTS2021 we also report the pixel-level AP and the best achievable Dice score over all thresholds, computed from the anomaly map \(A(x)\). All scores are averaged over \(K=8\) noise draws (Section 3.5). Every configuration is trained five times with different random seeds on the same split, and we report the mean and standard deviation.

MedIAnomaly provides no validation set, and choosing hyper-parameters on the test set would bias the comparison. We therefore report one common configuration for all datasets in the main comparison: \(\varepsilon=10\), input corruption and sparsity penalty on, which gave the best mean AUROC over the seven datasets. This single choice for all datasets limits, but does not remove, the optimism of selecting on test data; Section 5 shows that the mean AUROC changes by at most 0.012 for \(\varepsilon\) between 3 and 30. Results at the best \(\varepsilon\) for each dataset are given separately and are marked as an optimistic upper bound. The full \(\varepsilon\) sweep is shown in Section 5.

For other methods we quote the results reported by Cai et al.~(2025) and by Kar et al.~(2026, MSDE). These were obtained at the benchmark's input size, which may differ from ours, and in several cases with dataset-specific settings.

\hypertarget{results}{%
\section{Results}\label{results}}

\hypertarget{comparison-with-existing-methods}{%
\subsection{Comparison with existing methods}\label{comparison-with-existing-methods}}

Table 4 shows QBAE in the common setting, i.e.~one network and one configuration for all seven datasets. Its mean AUROC is 0.828. This is the highest of all methods in the MedIAnomaly benchmark; the best of these, AE-PL, reaches 0.820. MSDE (0.826) is on par with QBAE. However, MSDE chooses its backbone per dataset and refines the features with the test images, so the two results were not obtained under the same conditions. The difference of 0.002 is also smaller than the seed-to-seed variation of our mean (SD \ensuremath{\approx} 0.004). QBAE is clearly best on BraTS2021 (0.911, vs.~0.859 for DAE and 0.736 for MSDE). On ISIC2018 (0.751) it exceeds AE-PL, DAE and MSDE, but not the best self-supervised method (0.807). It is weakest on VinDr-CXR and Camelyon16, where methods based on pretrained feature densities perform better. AE-PL uses the same perceptual anomaly score as QBAE, and QBAE improves on it on four of the seven datasets. The largest gains are on ISIC2018 (+0.067) and BraTS2021 (+0.060). Note that our input resolution (\(128\times128\)) is higher than that of the benchmark (\(64\times64\)). On BraTS2021, the pixel-level AP is 0.838 and the best Dice score 0.773, compared with 0.755 and 0.711 for DAE (Table 5). Fig. 2 shows examples from the run with the median Dice score. The tumors are removed from the reconstructions, and the anomaly map follows the lesions. The examples include two difficult cases. The third image labeled normal contains white-matter hyperintensities consistent with age-related change; they are flagged, which is correct for an anomaly detector but counts as a false positive under the tumor labels. The last abnormal image has small, multifocal lesions, which give only a weak response.

\textbf{Table 4.} Image-level results in the common setting (one configuration for all datasets). AE-PL, DAE and the per-dataset best: MedIAnomaly (Cai et al., 2025, Table 6; AUROC, 64\ensuremath{\times}64 input, mean of repeated runs). MSDE: Kar et al.~(2026), Table 3 (fixed hyper-parameters; AUROC with AP in parentheses). Methods: AE-PL (Shvetsova et al., 2021), DAE (Kascenas et al., 2022), FAE-SSIM (Meissen et al., 2023), AE-U (Mao et al., 2020), MSC (Reiss and Hoshen, 2023), AnatPaste (Sato et al., 2023), AutoDDPM (Bercea et al., 2023). AE-PL uses the same perceptual term as our image score; DAE uses the input corruption that ours modifies. \textsuperscript{\dag}ImageNet pre-trained backbone. The last column picks the best method separately for each dataset; no single method attains all of these values. Among the methods in MedIAnomaly, AE-PL has the highest mean AUROC over the seven datasets (0.820).

\begin{longtable}[]{@{}
  l
  l
  l
  l
  l
  l
  l@{}}
\toprule\noalign{}
Dataset & \begin{tabular}[b]{@{}l@{}}QBAE\\AUROC\end{tabular} & \begin{tabular}[b]{@{}l@{}}QBAE\\AP\end{tabular} & AE-PL & DAE & \begin{tabular}[b]{@{}l@{}}MSDE\\AUROC (AP)\end{tabular} & \begin{tabular}[b]{@{}l@{}}Best in\\MedIAnomaly\end{tabular} \\
\midrule\noalign{}
\endhead
\bottomrule\noalign{}
\endlastfoot
RSNA & 0.881 \ensuremath{\pm} 0.002 & 0.858 \ensuremath{\pm} 0.005 & 0.875 & 0.861 & 0.918 (0.906) & 0.911 (FAE-SSIM\textsuperscript{\dag}) \\
VinDr-CXR & 0.723 \ensuremath{\pm} 0.009 & 0.730 \ensuremath{\pm} 0.010 & 0.753 & 0.686 & 0.819 (0.797) & 0.769 (AE-U) \\
Brain Tumor & 0.959 \ensuremath{\pm} 0.003 & 0.920 \ensuremath{\pm} 0.006 & 0.957 & 0.832 & 0.981 (0.981) & 0.973 (MSC\textsuperscript{\dag}) \\
LAG & 0.844 \ensuremath{\pm} 0.010 & 0.773 \ensuremath{\pm} 0.011 & 0.856 & 0.716 & 0.810 (0.831) & 0.856 (AE-PL\textsuperscript{\dag}) \\
ISIC2018 & 0.751 \ensuremath{\pm} 0.009 & 0.656 \ensuremath{\pm} 0.009 & 0.684 & 0.700 & 0.705 (0.638) & 0.807 (AnatPaste) \\
Camelyon16 & 0.726 \ensuremath{\pm} 0.005 & 0.647 \ensuremath{\pm} 0.006 & 0.761 & 0.654 & 0.812 (0.820) & 0.807 (AutoDDPM) \\
BraTS2021 & 0.911 \ensuremath{\pm} 0.017 & 0.956 \ensuremath{\pm} 0.010 & 0.851 & 0.859 & 0.736 (0.867) & 0.859 (DAE) \\
\textbf{Mean} & \textbf{0.828} & \textbf{0.792} & 0.820 & 0.758 & 0.826 (0.834) & -- \\
\end{longtable}

MSDE scores ImageNet- or AnatPaste-pretrained features (backbone chosen per dataset), refines them by mean shift over the union of the training and test images, and reports a single run.

\textbf{Table 5.} BraTS2021: image- and pixel-level results. Pixel AP and best Dice are computed from the map A(x) averaged over 8 noise draws.

\begin{longtable}[]{@{}
  l
  l
  l
  l@{}}
\toprule\noalign{}
Setting & Image AUROC & Pixel AP & Best Dice \\
\midrule\noalign{}
\endhead
\bottomrule\noalign{}
\endlastfoot
Common setting (\ensuremath{\varepsilon} = 10, KL on) & 0.911 \ensuremath{\pm} 0.017 & 0.838 \ensuremath{\pm} 0.016 & 0.773 \ensuremath{\pm} 0.017 \\
\ensuremath{\varepsilon} = 30, KL on & 0.939 \ensuremath{\pm} 0.003 & 0.849 \ensuremath{\pm} 0.008 & 0.786 \ensuremath{\pm} 0.006 \\
\ensuremath{\varepsilon} = 30, KL off (best on BraTS) & 0.942 \ensuremath{\pm} 0.003 & 0.855 \ensuremath{\pm} 0.002 & 0.792 \ensuremath{\pm} 0.003 \\
DAE (best in MedIAnomaly) & 0.859 & 0.755 & 0.711 \\
\end{longtable}

\begin{figure}
\centering
\includegraphics[width=1\textwidth,height=\textheight]{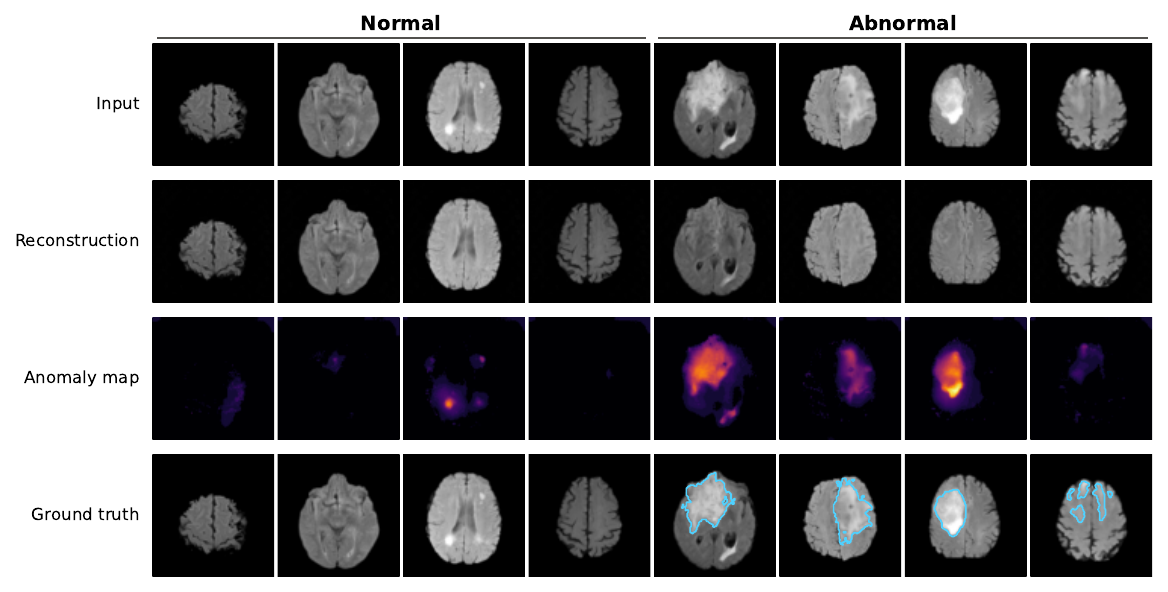}
\caption{\textbf{Fig. 2.} BraTS2021 examples from the common setting (run with the median Dice of five seeds). Left: four test images labeled normal; right: four abnormal images. Rows: input; reconstruction (one noise draw); anomaly map \(A(x)\) (shown from a noise-free pass of the same network; one square-root color scale for all eight images, without clipping); input with the ground-truth tumor outline.}
\end{figure}

\hypertarget{effect-of-the-information-budget}{%
\subsection{Effect of the information budget}\label{effect-of-the-information-budget}}

Fig. 3 shows how the AUROC depends on \(\varepsilon\). With input corruption, the mean AUROC changes by at most 0.012 for \(\varepsilon\) between 3 and 30 (0.816--0.828, about \(2\times10^4\) to \(1.2\times10^5\) bits). It drops for smaller budgets (0.751 at \(\varepsilon=0.3\)). Without the bottleneck (\(\varepsilon=10^8\)), the mean AUROC falls to 0.797. Without input corruption, the bottleneck is essential. The mean AUROC peaks at \(\varepsilon=1\) (0.805) and falls to 0.590 without the bottleneck. On BraTS2021 and LAG, the AUROC then drops below 0.5, i.e.~the network reconstructs abnormal images at least as well as normal ones. With corruption, the bottleneck has little effect on BraTS2021, Brain Tumor and VinDr-CXR, but adds 0.04--0.13 on Camelyon16, RSNA and LAG (Table 6). On BraTS2021, most of the improvement over the uncorrupted network thus comes from the input corruption.

The best budget depends on the dataset (Table 6). It is large (\(\varepsilon=30\)) for BraTS2021, Brain Tumor and VinDr-CXR, intermediate (\(\varepsilon=10\)) for LAG and ISIC2018, and small (\(\varepsilon=3\) and 1) for RSNA and Camelyon16. Choosing \(\varepsilon\) per dataset, with the architecture and all other settings fixed, raises the mean AUROC to 0.847. Because this choice is made on the test set, 0.847 is an optimistic upper bound. Still, it shows that the single parameter \(\varepsilon\), which sets the information budget, is enough to adapt one fixed network to datasets of very different kinds.

\begin{figure}
\centering
\includegraphics[width=1\textwidth,height=\textheight]{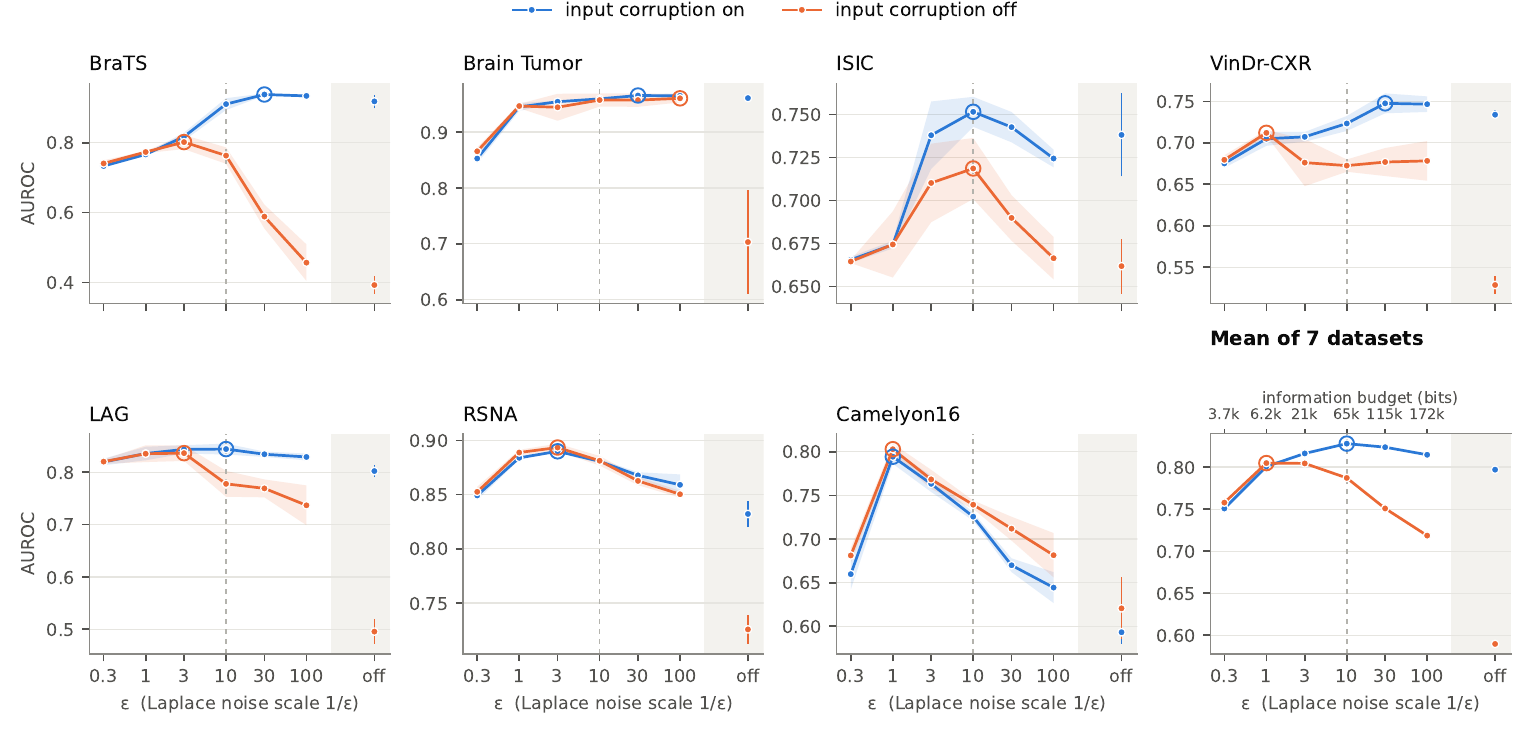}
\caption{\textbf{Fig. 3.} Image-level AUROC as a function of \ensuremath{\varepsilon} (KL on). Blue: input corruption on; orange: off. ``off'' = no bottleneck (\ensuremath{\varepsilon} = \(10^8\)). Shaded bands: \ensuremath{\pm}1 SD over 5 seeds; rings: best \ensuremath{\varepsilon} per curve; dashed line: common setting \ensuremath{\varepsilon} = 10. The top axis of the mean panel gives the information budget N\ensuremath{\cdot}c(\ensuremath{\varepsilon}).}
\end{figure}

\textbf{Table 6.} Effect of the QB bottleneck and of choosing \ensuremath{\varepsilon} per dataset (AUROC; corruption and KL on unless stated). ``QB off'' = \ensuremath{\varepsilon} = \(10^8\). ``Best \ensuremath{\varepsilon}'': only \ensuremath{\varepsilon} chosen per dataset. ``Best of all settings'': \ensuremath{\varepsilon}, corruption and KL all chosen. Both ``best'' columns are chosen on the test set and are optimistic upper bounds. Results without input corruption are shown in Fig. 3.

\begin{longtable}[]{@{}
  l
  l
  l
  l
  l@{}}
\toprule\noalign{}
Dataset & QB off & \ensuremath{\varepsilon} = 10 (common) & Best \ensuremath{\varepsilon} & Best of all settings \\
\midrule\noalign{}
\endhead
\bottomrule\noalign{}
\endlastfoot
RSNA & 0.832 & 0.881 & 0.890 (\ensuremath{\varepsilon} = 3) & 0.893 (\ensuremath{\varepsilon} = 3, corr. off, KL on) \\
VinDr-CXR & 0.734 & 0.723 & 0.748 (\ensuremath{\varepsilon} = 30) & 0.753 (\ensuremath{\varepsilon} = 100, corr. on, KL off) \\
Brain Tumor & 0.961 & 0.959 & 0.965 (\ensuremath{\varepsilon} = 30) & 0.968 (\ensuremath{\varepsilon} = off, corr. on, KL off) \\
LAG & 0.802 & 0.844 & 0.844 (\ensuremath{\varepsilon} = 10) & 0.860 (\ensuremath{\varepsilon} = 3, corr. on, KL off) \\
ISIC2018 & 0.738 & 0.751 & 0.751 (\ensuremath{\varepsilon} = 10) & 0.752 (\ensuremath{\varepsilon} = 30, corr. on, KL off) \\
Camelyon16 & 0.593 & 0.726 & 0.794 (\ensuremath{\varepsilon} = 1) & 0.803 (\ensuremath{\varepsilon} = 1, corr. off, KL on) \\
BraTS2021 & 0.919 & 0.911 & 0.939 (\ensuremath{\varepsilon} = 30) & 0.942 (\ensuremath{\varepsilon} = 30, corr. on, KL off) \\
\textbf{Mean} & 0.797 & 0.828 & 0.847 & 0.853 \\
\end{longtable}

\hypertarget{discussion}{%
\section{Discussion}\label{discussion}}

QBAE achieves three things. First, it uses a single architecture and a single configuration for all seven MedIAnomaly datasets and reaches a mean image-level AUROC of 0.828. To our knowledge, this is the highest mean AUROC reported for a method that does not adapt its architecture or settings to each dataset; the best previous such method is AE-PL, at 0.820. It is also on par with MSDE (0.826), which selects its feature extractor for each dataset and uses the test images. Second, it is the best reported method on BraTS2021, both at the image level (AUROC 0.911 vs.~0.859) and at the pixel level (AP 0.838 vs.~0.755). Third, it obtains these results with the noise kept at test time, so that every reconstruction is produced through QB layers whose information is provably bounded. A single parameter, \(\varepsilon\), sets this bound and adapts the same network to datasets of very different kinds.

\hypertarget{one-parameter-for-the-information-budget}{%
\subsection{One parameter for the information budget}\label{one-parameter-for-the-information-budget}}

In QBAE, the information that reaches the decoder is set by \(\varepsilon\) alone. Over the range we tested, the bound spans about \(4\times10^3\) to \(1.7\times10^5\) bits, and the network never changes. Other reconstruction-based methods tune the same quantity only indirectly, through the latent size, the depth or the skip connections. The mean AUROC is insensitive to the budget near its optimum: it varies by at most 0.012 for \(\varepsilon\) between 3 and 30, a factor of five in the budget. Hence a single default works across datasets. The best budget still depends on the dataset. It is large for BraTS2021, Brain Tumor and VinDr-CXR, and small for RSNA and Camelyon16. A plausible reading is that large, conspicuous lesions are best detected when the decoder receives enough detail to reconstruct normal tissue sharply, whereas subtle or diffuse changes require a tighter budget, so that they are not reconstructed. We did not test this interpretation directly.

\hypertarget{bottleneck-and-input-corruption-are-complementary}{%
\subsection{Bottleneck and input corruption are complementary}\label{bottleneck-and-input-corruption-are-complementary}}

The two mechanisms address the identity mapping in different ways. Without input corruption and without the bottleneck, the network copies its input: the mean AUROC falls to 0.590, and it drops below 0.5 on BraTS2021 and LAG. The bottleneck alone prevents this collapse and raises the mean to 0.805. With input corruption, the network can no longer simply copy, and the bottleneck adds less on average (0.797 to 0.828). Its gain, however, is concentrated where the corruption fits the anomalies least. On BraTS2021 and Brain Tumor, whose focal lesions resemble the blob-shaped corruption, the bottleneck adds little. On Camelyon16, RSNA and LAG, whose abnormalities are diffuse or subtle, it adds 0.04--0.13. The corruption thus tells the network \emph{what} to remove, and the budget limits \emph{how much} it can pass where the corruption gives no guidance.

\hypertarget{a-provable-budget-at-test-time}{%
\subsection{A provable budget at test time}\label{a-provable-budget-at-test-time}}

Because the noise is kept at test time, the network used for detection is the noisy network analyzed in Theorem 1, and every reported result is obtained under a stated information budget. Restricting the decoder to a provably limited view of the image did not prevent QBAE from reaching the results above. A budget stated in bits also opens uses beyond anomaly detection. For example, it bounds how much of an image could be recovered from what crosses the bottleneck, which may matter when encoder and decoder run on different sites.

\hypertarget{relation-to-feature-density-methods}{%
\subsection{Relation to feature-density methods}\label{relation-to-feature-density-methods}}

QBAE and MSDE reach the same mean AUROC, but their strengths differ. QBAE is clearly better on BraTS2021 and better on ISIC2018 and LAG. MSDE is better on VinDr-CXR, Camelyon16 and Brain Tumor. The two also rest on different resources. MSDE scores features of a pretrained network, chosen per dataset, and refines them with the test images. QBAE uses pretrained VGG-19 features only in its loss and its anomaly score; the representation that crosses the bottleneck is learned from normal training images alone. The two approaches are therefore complementary, and combining them is a natural next step.

\hypertarget{limitations}{%
\subsection{Limitations}\label{limitations}}

\textbf{Input corruption as a prior.} The blob-shaped corruption encodes a prior that anomalies are local and smooth. This prior matches large brain lesions well. On BraTS2021, most of the improvement comes from the corruption rather than from the bottleneck. Without corruption, the best AUROC is 0.802; with corruption but without the bottleneck, it is 0.919 (Fig. 3, Table 6). Part of our advantage on this dataset may therefore reflect how closely the synthetic perturbation resembles real tumors. The same criticism applies to DAE and to methods based on synthetic anomalies. The corruption is not specific to brain MRI, however. It uses random textures, does not model hyperintensity and is identical for all datasets. It also improves LAG, VinDr-CXR and ISIC2018 (by 0.03--0.07 at \(\varepsilon=10\)), while it does not help on RSNA and slightly lowers the AUROC on Camelyon16.

\textbf{Selection on the test set.} MedIAnomaly has no validation set. The common setting was chosen because it gave the best mean AUROC on the test sets. A single setting for all datasets and the insensitivity to \(\varepsilon\) limit this optimism, but do not remove it. The per-dataset results in Table 6 are upper bounds. Selecting \(\varepsilon\) without labels, for example from the reconstruction error on normal validation images, remains open.

\textbf{Scope of the bound.} Theorem 1 bounds the mutual information between the input and each reconstruction. It is an upper bound that need not be tight, and it does not by itself guarantee detection performance. It holds only because the noise is kept at test time. The averaged noisy score combines \(K=8\) independent passes; by Proposition 3, it can carry at most \(K\) times the per-reconstruction budget.

\textbf{Other.} QBAE depends on ImageNet-pretrained VGG-19 features for its loss and score. We use \(128\times128\) inputs, whereas the reported baselines use \(64\times64\); like the benchmark, we work on downsampled two-dimensional images rather than on full-resolution images or three-dimensional volumes. Pixel-level evaluation is possible only on BraTS2021. In BraTS2021, slices without tumor are labeled normal even when they show other abnormalities such as age-related white-matter hyperintensities (Fig. 2), so some detections count as false positives.

\hypertarget{conclusion}{%
\subsection{Conclusion}\label{conclusion}}

QBAE separates the information bottleneck of an autoencoder from its architecture. A quasi-binarizing layer on every encoder--decoder path bounds the information that reaches the decoder, and this bound depends only on \(\varepsilon\) and the number of QB elements. The encoder and decoder can therefore be as expressive as reconstruction requires. One network with one configuration reaches the highest mean AUROC on MedIAnomaly among methods that do not adapt to each dataset, and the best reported results on BraTS2021. Its single parameter \(\varepsilon\) adapts the network to datasets of very different kinds.

\hypertarget{credit-authorship-contribution-statement}{%
\section*{CRediT authorship contribution statement}\label{credit-authorship-contribution-statement}}
\addcontentsline{toc}{section}{CRediT authorship contribution statement}

\textbf{Shouhei Hanaoka:} Conceptualization, Methodology, Software, Formal analysis, Investigation, Visualization, Writing -- original draft, Funding acquisition. \textbf{Takahiro Nakao:} Validation, Writing -- review \& editing. \textbf{Atsushi Takamatsu:} Validation, Writing -- review \& editing. \textbf{Takeharu Yoshikawa:} Resources, Writing -- review \& editing. \textbf{Osamu Abe:} Supervision, Resources, Writing -- review \& editing.

\hypertarget{declaration-of-competing-interest}{%
\section*{Declaration of competing interest}\label{declaration-of-competing-interest}}
\addcontentsline{toc}{section}{Declaration of competing interest}

The Department of Computational Diagnostic Radiology and Preventive Medicine, The University of Tokyo Hospital, is sponsored by HIMEDIC Inc.~and Siemens Healthcare K.K. The sponsors had no role in the design of the study, the analysis, the writing of the manuscript or the decision to publish. The authors declare no other competing interests.

\hypertarget{declaration-of-generative-ai-and-ai-assisted-technologies-in-the-manuscript-preparation-process}{%
\section*{Declaration of generative AI and AI-assisted technologies in the manuscript preparation process}\label{declaration-of-generative-ai-and-ai-assisted-technologies-in-the-manuscript-preparation-process}}
\addcontentsline{toc}{section}{Declaration of generative AI and AI-assisted technologies in the manuscript preparation process}

During the preparation of this work the authors used Claude (Anthropic) in order to assist with implementing and debugging code, analyzing experimental logs, preparing figures and tables, and drafting and editing the text. After using this tool, the authors reviewed and edited the content as needed and take full responsibility for the content of the publication.

\hypertarget{ethics-statement}{%
\section*{Ethics statement}\label{ethics-statement}}
\addcontentsline{toc}{section}{Ethics statement}

This study used only publicly available, de-identified datasets (Section 4.1). No new data were collected, and no additional ethical approval was required.

\hypertarget{code-and-data-availability}{%
\section*{Code and data availability}\label{code-and-data-availability}}
\addcontentsline{toc}{section}{Code and data availability}

The source code of QBAE, including the training and evaluation scripts and the configurations used in this work, is available at \url{https://github.com/hanaokalog/MedIAnomalyQB}. All datasets are publicly available through the MedIAnomaly benchmark (Cai et al., 2025) and the original sources listed in Table 3.

\hypertarget{acknowledgments}{%
\section*{Acknowledgments}\label{acknowledgments}}
\addcontentsline{toc}{section}{Acknowledgments}

This work was supported by the Japan Society for the Promotion of Science (JSPS) KAKENHI Grant Number JP25K03479 and the Japan Science and Technology Agency (JST) CREST Grant Number JPMJCR21M2. The Department of Computational Diagnostic Radiology and Preventive Medicine, The University of Tokyo Hospital, is sponsored by HIMEDIC Inc.~and Siemens Healthcare K.K. This research was conducted using the Supermicro ARS-111GL-DNHR-LCC and FUJITSU Server PRIMERGY CX2550 M7 (Miyabi) at Joint Center for Advanced High Performance Computing (JCAHPC).

\clearpage

\appendix

\hypertarget{information-theoretic-analysis-of-the-qb-layer}{%
\section{Information-theoretic analysis of the QB layer}\label{information-theoretic-analysis-of-the-qb-layer}}

This appendix gives the formal statements behind Theorem 1 (Section 3.2.2) and their proofs.

\textbf{Setting.} Throughout, \(U\) denotes the encoder input together with any randomness used by the encoder; during training this is \(U=(X,\xi)\), with \(\xi\) the input corruption. Let \(z_i=g_i(U)=\sigma(h_i(U))\in[0,1]\) denote the pre-noise value of QB element \(i\), where \(g_i\) is a deterministic (measurable) function. The noisy output is \(\tilde Z_i = g_i(U)+N_i\). The noise variables \(N_1,\dots,N_N\) are i.i.d. \(\operatorname{Laplace}(0,b)\) with \(b=1/\varepsilon\), independent of \(U\). Natural logarithms are used, so information is measured in nats; to convert to bits, divide by \(\ln 2\).

\hypertarget{formal-statements}{%
\subsection{Formal statements}\label{formal-statements}}

\textbf{Proposition 1 (privacy bound).}
Each element \(\tilde Z_i\) is an \(\varepsilon\)-locally differentially private release of \(U\) (Duchi et al., 2013). Consequently,
\[
I(U;\tilde Z_i)\;\le\;\min\{\varepsilon,\varepsilon^2\}\ \text{nats}.
\tag{A.1}
\]

\textbf{Proposition 2 (amplitude bound).}
For every element,
\[
I(U;\tilde Z_i)\;\le\;C(\varepsilon)\;:=\;\tfrac12\ln\!\Big(2\pi e\big(\tfrac14+\tfrac{2}{\varepsilon^{2}}\big)\Big)-\ln\frac{2e}{\varepsilon}\ \text{nats}.
\tag{A.2}
\]

\textbf{Proposition 3 (additivity and data processing).}
Suppose \(\hat X\) depends on \(U\) only through \(\tilde Z=(\tilde Z_1,\dots,\tilde Z_N)\). Then
\[
I(X;\hat X)\;\le\; I(U;\tilde Z)\;\le\;\sum_{i=1}^{N} I(U;\tilde Z_i).
\tag{A.3}
\]

\textbf{Theorem 1 (formal).} Combining Propositions 1--3,
\[
I(X;\hat X)\;\le\;N\,c(\varepsilon),\qquad c(\varepsilon):=\min\big\{\varepsilon,\ \varepsilon^{2},\ C(\varepsilon)\big\}\ \text{nats}.
\tag{A.4}
\]
The privacy bound is the tighter one for \(\varepsilon\lesssim0.28\); the amplitude bound is tighter for all larger \(\varepsilon\), including every setting used in this work.

\hypertarget{proof-of-proposition-1}{%
\subsection{Proof of Proposition 1}\label{proof-of-proposition-1}}

The conditional density of \(\tilde Z_i\) given \(U=u\) is \(p(y\mid u)=\frac{1}{2b}\exp\!\big(-|y-g_i(u)|/b\big)\). For any two inputs \(u,u'\) and any \(y\in\mathbb{R}\), the triangle inequality gives
\[
\frac{p(y\mid u)}{p(y\mid u')}=\exp\!\Big(\frac{|y-g_i(u')|-|y-g_i(u)|}{b}\Big)\le \exp\!\Big(\frac{|g_i(u)-g_i(u')|}{b}\Big)\le e^{\varepsilon},
\tag{A.5}
\]
because \(g_i(u),g_i(u')\in[0,1]\) and \(1/b=\varepsilon\). Hence the mechanism \(u\mapsto\tilde Z_i\) is \(\varepsilon\)-differentially private with respect to arbitrary changes of its single input. This is \(\varepsilon\)-local differential privacy (Dwork et al., 2006; Duchi et al., 2013).

\textbf{Bound \(\varepsilon\).} Let \(p(y)=\mathbb{E}_{U'}[p(y\mid U')]\) be the marginal density of \(\tilde Z_i\). By (A.5), \(p(y)\ge e^{-\varepsilon}p(y\mid u)\) for every \(u\). Therefore
\[
I(U;\tilde Z_i)=\mathbb{E}\Big[\ln\frac{p(\tilde Z_i\mid U)}{p(\tilde Z_i)}\Big]\le \varepsilon .
\tag{A.6}
\]

\textbf{Bound \(\varepsilon^2\).} For \(\varepsilon<1\), the tighter bound \(I(U;\tilde Z_i)\le\varepsilon^2\) follows from Cuff and Yu (2016, Lemma 1), who show that \(\varepsilon\)-differential privacy implies mutual-information differential privacy with parameter \(\min\{\varepsilon,\varepsilon^2\}\) nats. Our setting is the special case of a single record. Combining the two bounds gives Proposition 1. \(\square\)

\hypertarget{proof-of-proposition-2}{%
\subsection{Proof of Proposition 2}\label{proof-of-proposition-2}}

\textbf{Reduction to \(Z_i\).} Since \(Z_i=g_i(U)\) and \(N_i\) is independent of \(U\), \(U\to Z_i\to\tilde Z_i\) is a Markov chain. By the data-processing inequality, \(I(U;\tilde Z_i)\le I(Z_i;\tilde Z_i)\). (In fact, equality holds.)

\textbf{Decomposition.} For an additive channel with noise independent of the input,
\[
I(Z_i;\tilde Z_i)=h(\tilde Z_i)-h(\tilde Z_i\mid Z_i)=h(\tilde Z_i)-h(N_i).
\]

\textbf{Noise entropy.} The differential entropy of \(\operatorname{Laplace}(0,b)\) is \(h(N_i)=1+\ln(2b)\).

\textbf{Output entropy.} Since \(Z_i\in[0,1]\), Popoviciu's inequality gives \(\operatorname{Var}(Z_i)\le\tfrac14\). Together with \(\operatorname{Var}(N_i)=2b^2\) and independence, \(\operatorname{Var}(\tilde Z_i)\le\tfrac14+2b^2\). Among all densities with a given variance, the Gaussian maximizes differential entropy (Cover and Thomas, 2006, Thm. 8.6.5). Hence
\[
h(\tilde Z_i)\le\tfrac12\ln\!\big(2\pi e(\tfrac14+2b^2)\big).
\]

\textbf{Conclusion.} Combining these and substituting \(b=1/\varepsilon\),
\[
I(U;\tilde Z_i)\le \tfrac12\ln\!\Big(2\pi e\big(\tfrac14+\tfrac{2}{\varepsilon^2}\big)\Big)-1-\ln\frac{2}{\varepsilon}=C(\varepsilon). \qquad\square
\]

\emph{Remark.} \(C(\varepsilon)\) is a (generally loose) upper bound on the capacity of the amplitude-constrained additive Laplace channel. It is used here only because it is explicit. For large \(\varepsilon\) it behaves as \(\ln\varepsilon+\tfrac12\ln(\pi/(8e))+o(1)\), i.e.~it grows logarithmically; as \(\varepsilon\to0\) it tends to the constant \(\tfrac12\ln(\pi/e)\approx0.072\) nats rather than to zero, which is why the privacy bound is the tighter one for small \(\varepsilon\).

\hypertarget{proof-of-proposition-3}{%
\subsection{Proof of Proposition 3}\label{proof-of-proposition-3}}

\textbf{First inequality.} By assumption, \(\hat X\) is a (possibly randomized) function of \(\tilde Z\) alone, with any randomness independent of everything else. Hence \(X\to U\to\tilde Z\to\hat X\) is a Markov chain, since \(X\) is a component of \(U\). Two applications of the data-processing inequality give
\[
I(X;\hat X)\le I(X;\tilde Z)\le I(U;\tilde Z).
\]

\textbf{Second inequality.} Conditionally on \(U\), the outputs \(\tilde Z_i=g_i(U)+N_i\) are independent, because the \(N_i\) are independent of each other and of \(U\). Hence \(h(\tilde Z\mid U)=\sum_i h(N_i)=\sum_i h(\tilde Z_i\mid U)\). By subadditivity of differential entropy, \(h(\tilde Z)\le\sum_i h(\tilde Z_i)\). Therefore
\[
I(U;\tilde Z)=h(\tilde Z)-h(\tilde Z\mid U)\le\sum_{i=1}^N\big[h(\tilde Z_i)-h(\tilde Z_i\mid U)\big]=\sum_{i=1}^N I(U;\tilde Z_i).
\]

\textbf{Third inequality.} Each summand is bounded by Propositions 1 and 2. \(\square\)

\emph{Remark.} The sum of the privacy bounds alone could also be obtained from the basic composition theorem of differential privacy: the joint release is \(N\varepsilon\)-DP, so (A.6) gives \(I(U;\tilde Z)\le N\varepsilon\). The argument above is needed to sum the amplitude bounds.

\hypertarget{numerical-values}{%
\subsection{Numerical values}\label{numerical-values}}

The per-element values in Table 1 are \(c(\varepsilon)/\ln 2\), i.e.~the smaller of \(\min\{\varepsilon,\varepsilon^2\}/\ln2\) (privacy) and \(C(\varepsilon)/\ln2\) (amplitude). The totals multiply this by \(N=32{,}768\).

\begin{longtable}[]{@{}
  l
  l
  l
  l
  l@{}}
\toprule\noalign{}
\(\varepsilon\) & \(\min\{\varepsilon,\varepsilon^2\}\) {[}nats{]} & \(C(\varepsilon)\) {[}nats{]} & Bound per element {[}bits{]} & Total {[}bits{]} \\
\midrule\noalign{}
\endhead
\bottomrule\noalign{}
\endlastfoot
0.3 & 0.09 & 0.078 & 0.112 & 3,685 \\
1 & 1 & 0.131 & 0.189 & 6,205 \\
3 & 3 & 0.449 & 0.648 & 21,238 \\
10 & 10 & 1.374 & 1.982 & 64,941 \\
30 & 30 & 2.438 & 3.518 & 115,267 \\
100 & 100 & 3.638 & 5.249 & 171,994 \\
\end{longtable}

At \(\varepsilon=0.3\) the two bounds are close (0.090 vs 0.078 nats), consistent with the crossover at \(\varepsilon\approx0.28\) stated after (A.4).

\hypertarget{network-details}{%
\section{Network details}\label{network-details}}

Fig. B.1 shows every layer of the network described in Section 3.3; Table B.1 lists the levels. The network has 60.7 M parameters. Most of them sit in the two deepest decoder levels (Table B.1), which operate on \(4\times4\) and \(8\times8\) feature maps and therefore add little computation; a forward pass costs 4.2 GFLOPs per \(128\times128\) image (2.1 G multiply--accumulates).

\begin{figure}
\centering
\includegraphics[width=1\textwidth,height=0.82\textheight]{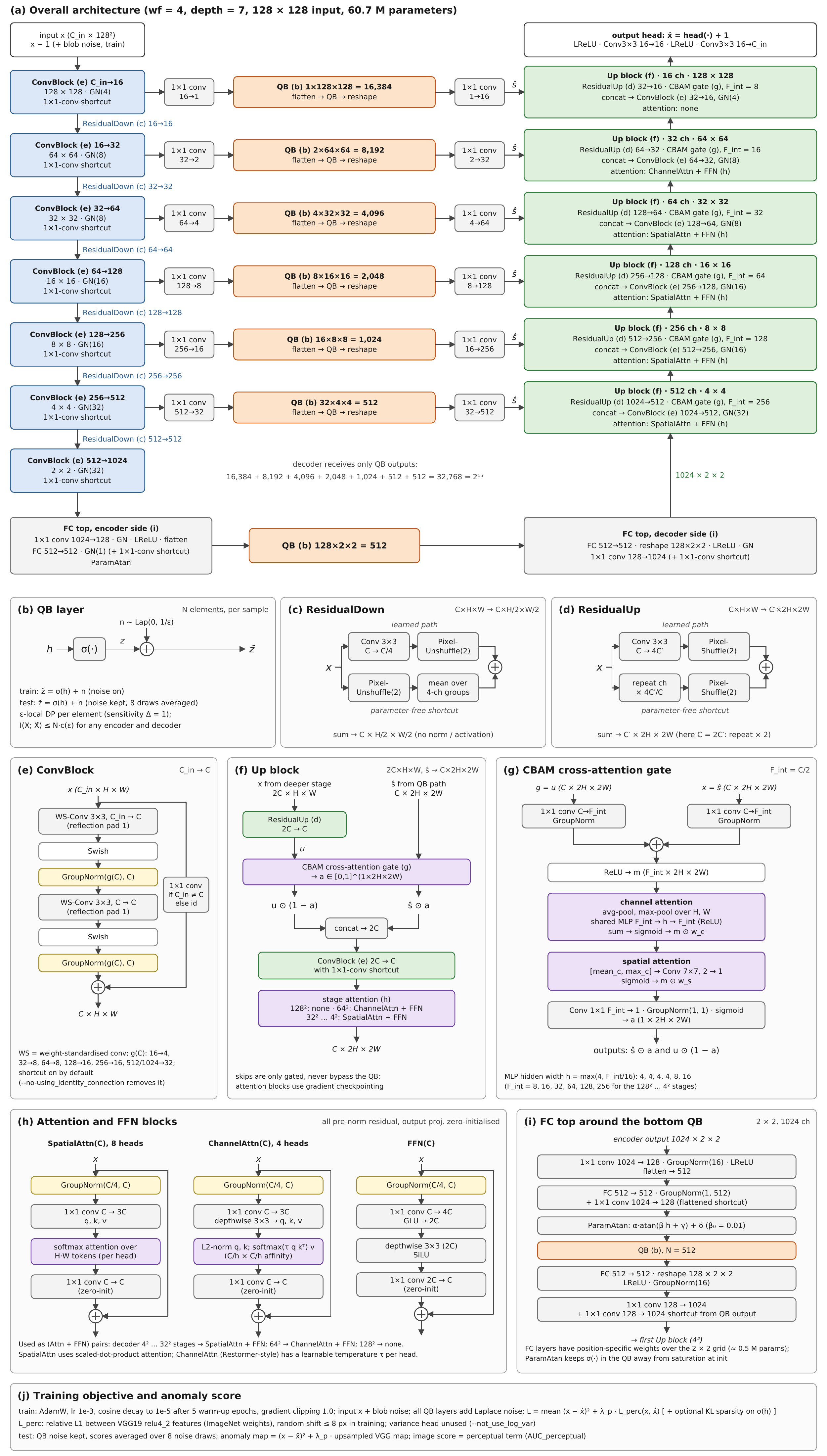}
\caption{\textbf{Fig. B.1.} Detailed architecture of QBAE. (a) Overall network. (b) QB layer. (c), (d) DC-AE residual down- and upsampling. (e) Convolution block. (f) Decoder (up) block. (g) Cross-attention gate. (h) Attention and feed-forward blocks. (i) Fully connected layers around the bottom QB layer. (j) Training objective and anomaly score.}
\end{figure}

\textbf{Table B.1.} Levels of the network (\(128\times128\) input).

\begin{longtable}[]{@{}
  l
  l
  l
  l
  l
  l@{}}
\toprule\noalign{}
Level & Resolution & Width & \begin{tabular}[b]{@{}l@{}}QB elements\\(channels per pixel)\end{tabular} & \begin{tabular}[b]{@{}l@{}}Decoder\\attention\end{tabular} & \begin{tabular}[b]{@{}l@{}}Parameters\\(enc. / dec.)\end{tabular} \\
\midrule\noalign{}
\endhead
\bottomrule\noalign{}
\endlastfoot
1 & \(128^2\) & 16 & 16,384 (1) & -- & 0.003 M / 0.03 M \\
2 & \(64^2\) & 32 & 8,192 (2) & channel attention + FFN & 0.015 M / 0.12 M \\
3 & \(32^2\) & 64 & 4,096 (4) & spatial attention + FFN & 0.06 M / 0.46 M \\
4 & \(16^2\) & 128 & 2,048 (8) & spatial attention + FFN & 0.23 M / 1.84 M \\
5 & \(8^2\) & 256 & 1,024 (16) & spatial attention + FFN & 0.92 M / 7.35 M \\
6 & \(4^2\) & 512 & 512 (32) & spatial attention + FFN & 3.67 M / 29.39 M \\
7 (bottom) & \(2^2\) & 1024 & 512 (fully connected) & -- & 14.69 M / -- \\
\end{longtable}

Not listed: DC-AE downsampling layers (0.79 M), \(1\times1\) convolutions of the skip connections (0.05 M), fully connected bottom layers (1.05 M) and output head (0.005 M). Decoder parameters include the upsampling layer, cross-attention gate, convolution block and attention block of each level.

\textbf{Components.} Convolution blocks use two weight-standardized (Qiao et al., 2019) \(3\times3\) convolutions with reflection padding, each followed by Swish and GroupNorm, and a residual connection (\(1\times1\) convolution when the width changes). DC-AE downsampling adds a learned path (convolution followed by pixel unshuffle; Shi et al., 2016) to a parameter-free path (pixel unshuffle followed by averaging over channel groups); upsampling mirrors this with pixel shuffle and channel duplication. The cross-attention gate projects the upsampled decoder features \(u\) and the post-QB skip features \(\hat s\) to \(F_{\mathrm{int}}\) channels (half the level width), applies CBAM (Woo et al., 2018) channel and spatial attention to their sum, and outputs a map \(a\in[0,1]\); the decoder then concatenates \(u\odot(1-a)\) and \(\hat s\odot a\). Spatial attention uses 8 heads of softmax self-attention over all positions; channel attention follows Restormer (Zamir et al., 2022; 4 heads, transposed attention); the feed-forward block is a gated \(1\times1\) / depth-wise \(3\times3\) network. All attention and feed-forward blocks are pre-normalized residual blocks whose output projections are initialized to zero. At the bottom, a \(1\times1\) convolution reduces the \(1024\times2\times2\) map to \(128\times2\times2\), which is flattened and passed through a fully connected layer \(512\to512\) with GroupNorm and a \(1\times1\)-convolution shortcut, the learnable arctangent of Section 3.2.3, the QB layer, and a mirrored fully connected layer back to \(1024\times2\times2\).

\hypertarget{input-corruption}{%
\section{Input corruption}\label{input-corruption}}

For an image \(x\in\mathbb{R}^{C\times H\times W}\), the corruption \(\eta\) of Section 3.4 is generated as follows. All random variables are drawn independently for each image and each training step.

\begin{enumerate}
\def\labelenumi{\arabic{enumi}.}
\tightlist
\item
  Draw the smoothing scales \(r_1,r_2\sim\mathcal{U}(1,17)\) pixels, a threshold \(t\sim\mathcal{U}(0,2)\) and an exponent \(p\sim\mathcal{U}(0.01,2.01)\).
\item
  Texture field: draw white Gaussian noise \(w_1\in\mathbb{R}^{C\times H\times W}\), convolve each channel with a Gaussian kernel of standard deviation \(r_1\), and normalize the result \(g_1\) to unit standard deviation over the image.
\item
  Mask field: draw \(w_2\in\mathbb{R}^{H\times W}\), smooth it with standard deviation \(r_2\) and normalize it to \(g_2\) in the same way. The blob mask is
  \[
  m=\max(g_2-t,\,0)^{\,p},
  \]
  which is zero outside a few smooth patches; larger \(t\) gives fewer and smaller blobs, and \(p\) controls how sharp their edges are.
\item
  The corruption is
  \[
  \eta=\sigma\,\mathrm{std}(x)\;g_1\odot m,
  \]
  with the mask shared by all channels, \(\mathrm{std}(x)\) the standard deviation of the image and \(\sigma=1\) in all experiments.
\end{enumerate}

Compared with the coarse noise of Kascenas et al.~(2022), which perturbs the whole image with upsampled low-resolution Gaussian noise, this corruption alters only parts of the image, at scales that vary from image to image, and leaves the rest unchanged. In our implementation the Gaussian smoothing is done on the GPU in the Fourier domain, on a canvas of twice the image size that is then cropped, so that the periodic boundary of the FFT has no visible effect.

\hypertarget{other-test-time-readouts}{%
\section{Other test-time readouts}\label{other-test-time-readouts}}

Instead of keeping the noise, the QB layers can be read out without noise (\(\tilde z_i=z_i\)) or binarized (\(\tilde z_i=\mathbb{1}[z_i>\tfrac12]\)). Theorem 1 does not cover the noise-free readout; the binarized readout limits the information to \(N\) bits, since each element takes only two values. In the common setting, the mean AUROC over the seven datasets is 0.817 without noise and 0.812 with binarization, compared with 0.828 with the noise kept and averaged over \(K=8\) draws. Binarization costs little except on BraTS2021 (0.852 vs.~0.911), which indicates that the learned code is close to binary. The information carried by the binarized code, estimated as \(\sum_i h_2(p_i)\) from the firing frequencies \(p_i\) of the elements on the test set (\(h_2\): binary entropy), is on average 14,800 bits per image with the sparsity penalty and 26,300 bits without it.

\clearpage

\section*{References}
\begin{small}
\begin{list}{}{\setlength{\leftmargin}{1.5em}\setlength{\itemindent}{-1.5em}\setlength{\itemsep}{2pt}}
\item \addcontentsline{toc}{section}{References}
\item Abadi, M., Chu, A., Goodfellow, I., McMahan, H.B., Mironov, I., Talwar, K., Zhang, L., 2016. Deep learning with differential privacy, in: Proceedings of the 2016 ACM SIGSAC Conference on Computer and Communications Security (CCS), pp.~308--318.
\item Alemi, A.A., Fischer, I., Dillon, J.V., Murphy, K., 2017. Deep variational information bottleneck, in: International Conference on Learning Representations (ICLR).
\item Alemi, A., Poole, B., Fischer, I., Dillon, J., Saurous, R.A., Murphy, K., 2018. Fixing a broken ELBO, in: Proceedings of the 35th International Conference on Machine Learning (ICML), PMLR 80, pp.~159--168.
\item Baid, U., Ghodasara, S., Mohan, S., Bilello, M., Calabrese, E., Colak, E., Farahani, K., Kalpathy-Cramer, J., Kitamura, F.C., Pati, S., Prevedello, L.M., Rudie, J.D., Sako, C., Shinohara, R.T., Bergquist, T., Chai, R., Eddy, J., Elliott, J., Reade, W., Schaffter, T., Yu, T., Zheng, J., Moawad, A.W., Coelho, L.O., McDonnell, O., Miller, E., Moron, F.E., Oswood, M.C., Shih, R.Y., Siakallis, L., Bronstein, Y., Mason, J.R., Miller, A.F., Choudhary, G., Agarwal, A., Besada, C.H., Derakhshan, J.J., Diogo, M.C., Do-Dai, D.D., Farage, L., Go, J.L., Hadi, M., Hill, V.B., Iv, M., Joyner, D., Lincoln, C., Lotan, E., Miyakoshi, A., Sanchez-Montano, M., Nath, J., Nguyen, X.V., Nicolas-Jilwan, M., Jimenez, J.O., Ozturk, K., Petrovic, B.D., Shah, C., Shah, L.M., Sharma, M., Simsek, O., Singh, A.K., Soman, S., Statsevych, V., Weinberg, B.D., Young, R.J., Ikuta, I., Agarwal, A.K., Cambron, S.C., Silbergleit, R., Dusoi, A., Postma, A.A., Letourneau-Guillon, L., Guzman Perez-Carrillo, G.J., Saha, A., Soni, N., Zaharchuk, G., Zohrabian, V.M., Chen, Y., Cekic, M.M., Rahman, A., Small, J.E., Sethi, V., Davatzikos, C., Mongan, J., Hess, C., Cha, S., Villanueva-Meyer, J., Freymann, J.B., Kirby, J.S., Wiestler, B., Crivellaro, P., Colen, R.R., Kotrotsou, A., Marcus, D., Milchenko, M., Nazeri, A., Fathallah-Shaykh, H., Wiest, R., Jakab, A., Weber, M.-A., Mahajan, A., Menze, B., Flanders, A.E., Bakas, S., 2021. The RSNA-ASNR-MICCAI BraTS 2021 benchmark on brain tumor segmentation and radiogenomic classification. arXiv:2107.02314.
\item Ballé, J., Laparra, V., Simoncelli, E.P., 2017. End-to-end optimized image compression, in: International Conference on Learning Representations (ICLR).
\item Bao, J., Sun, H., Deng, H., He, Y., Zhang, Z., Li, X., 2024. BMAD: Benchmarks for medical anomaly detection, in: Proceedings of the IEEE/CVF Conference on Computer Vision and Pattern Recognition Workshops (CVPRW), pp.~4042--4053.
\item Baur, C., Denner, S., Wiestler, B., Navab, N., Albarqouni, S., 2021. Autoencoders for unsupervised anomaly segmentation in brain MR images: A comparative study. Medical Image Analysis 69, 101952. \url{https://doi.org/10.1016/j.media.2020.101952}
\item Bercea, C.I., Neumayr, M., Rueckert, D., Schnabel, J.A., 2023. Mask, stitch, and re-sample: Enhancing robustness and generalizability in anomaly detection through automatic diffusion models, in: ICML 2023 Workshop on Interpretable Machine Learning in Healthcare. arXiv:2305.19643.
\item Cai, Y., Chen, H., Cheng, K.-T., 2024. Rethinking autoencoders for medical anomaly detection from a theoretical perspective, in: Medical Image Computing and Computer Assisted Intervention (MICCAI), LNCS 15011, pp.~544--554. \url{https://doi.org/10.1007/978-3-031-72120-5_51}
\item Cai, Y., Zhang, W., Chen, H., Cheng, K.-T., 2025. MedIAnomaly: A comparative study of anomaly detection in medical images. Medical Image Analysis 102, 103500. \url{https://doi.org/10.1016/j.media.2025.103500}
\item Chen, J., Cai, H., Chen, J., Xie, E., Yang, S., Tang, H., Li, M., Lu, Y., Han, S., 2025. Deep compression autoencoder for efficient high-resolution diffusion models, in: International Conference on Learning Representations (ICLR).
\item Cheng, J., Huang, W., Cao, S., Yang, R., Yang, W., Yun, Z., Wang, Z., Feng, Q., 2015. Enhanced performance of brain tumor classification via tumor region augmentation and partition. PLoS ONE 10, e0140381.
\item Choi, K., Tatwawadi, K., Grover, A., Weissman, T., Ermon, S., 2019. Neural joint source-channel coding, in: Proceedings of the 36th International Conference on Machine Learning (ICML), PMLR 97, pp.~1182--1192.
\item Codella, N., Rotemberg, V., Tschandl, P., Celebi, M.E., Dusza, S., Gutman, D., Helba, B., Kalloo, A., Liopyris, K., Marchetti, M., Kittler, H., Halpern, A., 2019. Skin lesion analysis toward melanoma detection 2018: A challenge hosted by the International Skin Imaging Collaboration (ISIC). arXiv:1902.03368.
\item Cover, T.M., Thomas, J.A., 2006. Elements of Information Theory, 2nd ed.~Wiley, Hoboken, NJ.
\item Cuff, P., Yu, L., 2016. Differential privacy as a mutual information constraint, in: Proceedings of the 2016 ACM SIGSAC Conference on Computer and Communications Security (CCS), pp.~43--54. \url{https://doi.org/10.1145/2976749.2978308}
\item Deng, J., Dong, W., Socher, R., Li, L.-J., Li, K., Fei-Fei, L., 2009. ImageNet: A large-scale hierarchical image database, in: IEEE Conference on Computer Vision and Pattern Recognition (CVPR), pp.~248--255.
\item Duchi, J.C., Jordan, M.I., Wainwright, M.J., 2013. Local privacy and statistical minimax rates, in: IEEE 54th Annual Symposium on Foundations of Computer Science (FOCS), pp.~429--438. \url{https://doi.org/10.1109/FOCS.2013.53}
\item Dwork, C., McSherry, F., Nissim, K., Smith, A., 2006. Calibrating noise to sensitivity in private data analysis, in: Theory of Cryptography Conference (TCC), LNCS 3876, pp.~265--284.
\item Ehteshami Bejnordi, B., Veta, M., van Diest, P.J., van Ginneken, B., Karssemeijer, N., Litjens, G., van der Laak, J.A.W.M., the CAMELYON16 Consortium, Hermsen, M., Manson, Q.F., Balkenhol, M., Geessink, O., Stathonikos, N., van Dijk, M.C.R.F., Bult, P., Beca, F., Beck, A.H., Wang, D., Khosla, A., Gargeya, R., Irshad, H., Zhong, A., Dou, Q., Li, Q., Chen, H., Lin, H.-J., Heng, P.-A., Haß, C., Bruni, E., Wong, Q., Halici, U., {\"O}ner, M.{\"U}., Cetin-Atalay, R., Berseth, M., Khvatkov, V., Vylegzhanin, A., Kraus, O., Shaban, M., Rajpoot, N., Awan, R., Sirinukunwattana, K., Qaiser, T., Tsang, Y.-W., Tellez, D., Annuscheit, J., Hufnagl, P., Valkonen, M., Kartasalo, K., Latonen, L., Ruusuvuori, P., Liimatainen, K., Albarqouni, S., Mungal, B., George, A., Demirci, S., Navab, N., Watanabe, S., Seno, S., Takenaka, Y., Matsuda, H., Ahmady Phoulady, H., Kovalev, V., Kalinovsky, A., Liauchuk, V., Bueno, G., Fernandez-Carrobles, M.M., Serrano, I., Deniz, O., Racoceanu, D., Ven{\^a}ncio, R., 2017. Diagnostic assessment of deep learning algorithms for detection of lymph node metastases in women with breast cancer. JAMA 318, 2199--2210.
\item Gong, D., Liu, L., Le, V., Saha, B., Mansour, M.R., Venkatesh, S., van den Hengel, A., 2019. Memorizing normality to detect anomaly: Memory-augmented deep autoencoder for unsupervised anomaly detection, in: IEEE/CVF International Conference on Computer Vision (ICCV), pp.~1705--1714.
\item Hamada, A., 2025. Br35H :: Brain tumor detection 2020. IEEE Dataport. \url{https://doi.org/10.21227/tbkk-q937}
\item Higgins, I., Matthey, L., Pal, A., Burgess, C., Glorot, X., Botvinick, M., Mohamed, S., Lerchner, A., 2017. \ensuremath{\beta}-VAE: Learning basic visual concepts with a constrained variational framework, in: International Conference on Learning Representations (ICLR).
\item Johnson, J., Alahi, A., Fei-Fei, L., 2016. Perceptual losses for real-time style transfer and super-resolution, in: European Conference on Computer Vision (ECCV), LNCS 9906, pp.~694--711.
\item Kaiser, Ł., Bengio, S., 2018. Discrete autoencoders for sequence models. arXiv:1801.09797.
\item Kar, P., Lakshmi S, G., Bej, S., 2026. Improved anomaly detection in medical images via mean shift density enhancement. arXiv:2604.19191v1.
\item Kascenas, A., Pugeault, N., O'Neil, A.Q., 2022. Denoising autoencoders for unsupervised anomaly detection in brain MRI, in: Proceedings of the 5th International Conference on Medical Imaging with Deep Learning (MIDL), PMLR 172, pp.~653--664.
\item Kascenas, A., Sanchez, P., Schrempf, P., Wang, C., Clackett, W., Mikhael, S.S., Voisey, J.P., Goatman, K., Weir, A., Pugeault, N., Tsaftaris, S.A., O'Neil, A.Q., 2023. The role of noise in denoising models for anomaly detection in medical images. Medical Image Analysis 90, 102963. \url{https://doi.org/10.1016/j.media.2023.102963}
\item Kingma, D.P., Welling, M., 2014. Auto-encoding variational Bayes, in: International Conference on Learning Representations (ICLR).
\item Li, L., Xu, M., Wang, X., Jiang, L., Liu, H., 2019. Attention based glaucoma detection: A large-scale database and CNN model, in: IEEE/CVF Conference on Computer Vision and Pattern Recognition (CVPR), pp.~10571--10580.
\item Loshchilov, I., Hutter, F., 2017. SGDR: Stochastic gradient descent with warm restarts, in: International Conference on Learning Representations (ICLR).
\item Loshchilov, I., Hutter, F., 2019. Decoupled weight decay regularization, in: International Conference on Learning Representations (ICLR).
\item Mao, Y., Xue, F.-F., Wang, R., Zhang, J., Zheng, W.-S., Liu, H., 2020. Abnormality detection in chest X-ray images using uncertainty prediction autoencoders, in: Medical Image Computing and Computer Assisted Intervention (MICCAI), LNCS 12266, pp.~529--538. \url{https://doi.org/10.1007/978-3-030-59725-2_51}
\item McGregor, A., Mironov, I., Pitassi, T., Reingold, O., Talwar, K., Vadhan, S., 2010. The limits of two-party differential privacy, in: IEEE 51st Annual Symposium on Foundations of Computer Science (FOCS), pp.~81--90.
\item Meissen, F., Paetzold, J., Kaissis, G., Rueckert, D., 2023. Unsupervised anomaly localization with structural feature-autoencoders, in: Brainlesion: Glioma, Multiple Sclerosis, Stroke and Traumatic Brain Injuries (BrainLes 2022), LNCS 13769, pp.~14--24. \url{https://doi.org/10.1007/978-3-031-33842-7_2}
\item Mireshghallah, F., Taram, M., Ramrakhyani, P., Jalali, A., Tullsen, D., Esmaeilzadeh, H., 2020. Shredder: Learning noise distributions to protect inference privacy, in: Proceedings of the 25th International Conference on Architectural Support for Programming Languages and Operating Systems (ASPLOS). \url{https://doi.org/10.1145/3373376.3378522}
\item Ng, A., 2011. Sparse autoencoder. CS294A Lecture notes, Stanford University.
\item Nguyen, H.Q., Lam, K., Le, L.T., Pham, H.H., Tran, D.Q., Nguyen, D.B., Le, D.D., Pham, C.M., Tong, H.T.T., Dinh, D.H., Do, C.D., Doan, L.T., Nguyen, C.N., Nguyen, B.T., Nguyen, Q.V., Hoang, A.D., Phan, H.N., Nguyen, A.T., Ho, P.H., Ngo, D.T., Nguyen, N.T., Nguyen, N.T., Dao, M., Vu, V., 2022. VinDr-CXR: An open dataset of chest X-rays with radiologist's annotations. Scientific Data 9, 429.
\item Oktay, O., Schlemper, J., Le Folgoc, L., Lee, M., Heinrich, M., Misawa, K., Mori, K., McDonagh, S., Hammerla, N.Y., Kainz, B., Glocker, B., Rueckert, D., 2018. Attention U-Net: Learning where to look for the pancreas, in: Medical Imaging with Deep Learning (MIDL). arXiv:1804.03999.
\item Pascanu, R., Mikolov, T., Bengio, Y., 2013. On the difficulty of training recurrent neural networks, in: Proceedings of the 30th International Conference on Machine Learning (ICML), PMLR 28, pp.~1310--1318.
\item Pinaya, W.H.L., Graham, M.S., Gray, R., da Costa, P.F., Tudosiu, P.-D., Wright, P., Mah, Y.H., MacKinnon, A.D., Teo, J.T., Jager, R., Werring, D., Rees, G., Nachev, P., Ourselin, S., Cardoso, M.J., 2022. Fast unsupervised brain anomaly detection and segmentation with diffusion models, in: Medical Image Computing and Computer Assisted Intervention (MICCAI), pp.~705--714. \url{https://doi.org/10.1007/978-3-031-16452-1_67}
\item Qiao, S., Wang, H., Liu, C., Shen, W., Yuille, A., 2019. Micro-batch training with batch-channel normalization and weight standardization. arXiv:1903.10520.
\item Ramachandran, P., Zoph, B., Le, Q.V., 2017. Searching for activation functions. arXiv:1710.05941.
\item Reiss, T., Hoshen, Y., 2023. Mean-shifted contrastive loss for anomaly detection, in: Proceedings of the AAAI Conference on Artificial Intelligence 37(2), pp.~2155--2162. \url{https://doi.org/10.1609/aaai.v37i2.25309}
\item Ronneberger, O., Fischer, P., Brox, T., 2015. U-Net: Convolutional networks for biomedical image segmentation, in: Medical Image Computing and Computer-Assisted Intervention (MICCAI), LNCS 9351, pp.~234--241.
\item Salakhutdinov, R., Hinton, G., 2009. Semantic hashing. International Journal of Approximate Reasoning 50, 969--978.
\item Saleh, A., Sukaik, R., Abu-Naser, S.S., 2020. Brain tumor classification using deep learning, in: International Conference on Assistive and Rehabilitation Technologies (iCareTech), IEEE, pp.~131--136.
\item Sato, J., Suzuki, Y., Wataya, T., Nishigaki, D., Kita, K., Yamagata, K., Tomiyama, N., Kido, S., 2023. Anatomy-aware self-supervised learning for anomaly detection in chest radiographs. iScience 26, 107086. \url{https://doi.org/10.1016/j.isci.2023.107086}
\item Schlegl, T., Seeb{\"o}ck, P., Waldstein, S.M., Langs, G., Schmidt-Erfurth, U., 2019. f-AnoGAN: Fast unsupervised anomaly detection with generative adversarial networks. Medical Image Analysis 54, 30--44.
\item Shi, W., Caballero, J., Husz{\'a}r, F., Totz, J., Aitken, A.P., Bishop, R., Rueckert, D., Wang, Z., 2016. Real-time single image and video super-resolution using an efficient sub-pixel convolutional neural network, in: IEEE Conference on Computer Vision and Pattern Recognition (CVPR), pp.~1874--1883.
\item Shih, G., Wu, C.C., Halabi, S.S., Kohli, M.D., Prevedello, L.M., Cook, T.S., Sharma, A., Amorosa, J.K., Arteaga, V., Galperin-Aizenberg, M., Gill, R.R., Godoy, M.C.B., Hobbs, S., Jeudy, J., Laroia, A., Shah, P.N., Vummidi, D., Yaddanapudi, K., Stein, A., 2019. Augmenting the National Institutes of Health chest radiograph dataset with expert annotations of possible pneumonia. Radiology: Artificial Intelligence 1, e180041. \url{https://doi.org/10.1148/ryai.2019180041}
\item Shvetsova, N., Bakker, B., Fedulova, I., Schulz, H., Dylov, D.V., 2021. Anomaly detection in medical imaging with deep perceptual autoencoders. IEEE Access 9, 118571--118583. \url{https://doi.org/10.1109/ACCESS.2021.3107163}
\item Simonyan, K., Zisserman, A., 2015. Very deep convolutional networks for large-scale image recognition, in: International Conference on Learning Representations (ICLR).
\item Tishby, N., Pereira, F.C., Bialek, W., 1999. The information bottleneck method, in: Proceedings of the 37th Annual Allerton Conference on Communication, Control and Computing, pp.~368--377.
\item van den Oord, A., Vinyals, O., Kavukcuoglu, K., 2017. Neural discrete representation learning, in: Advances in Neural Information Processing Systems (NeurIPS) 30.
\item Vincent, P., Larochelle, H., Bengio, Y., Manzagol, P.-A., 2008. Extracting and composing robust features with denoising autoencoders, in: Proceedings of the 25th International Conference on Machine Learning (ICML), pp.~1096--1103.
\item Wolleb, J., Bieder, F., Sandk{\"u}hler, R., Cattin, P.C., 2022. Diffusion models for medical anomaly detection, in: Medical Image Computing and Computer Assisted Intervention (MICCAI), LNCS 13438, pp.~35--45. \url{https://doi.org/10.1007/978-3-031-16452-1_4}
\item Woo, S., Park, J., Lee, J.-Y., Kweon, I.S., 2018. CBAM: Convolutional block attention module, in: European Conference on Computer Vision (ECCV), LNCS 11211, pp.~3--19.
\item Wu, Y., He, K., 2018. Group normalization, in: European Conference on Computer Vision (ECCV), LNCS 11217, pp.~3--19.
\item Wyatt, J., Leach, A., Schmon, S.M., Willcocks, C.G., 2022. AnoDDPM: Anomaly detection with denoising diffusion probabilistic models using simplex noise, in: IEEE/CVF Conference on Computer Vision and Pattern Recognition Workshops (CVPRW), pp.~650--656.
\item You, Z., Cui, L., Shen, Y., Yang, K., Lu, X., Zheng, Y., Le, X., 2022. A unified model for multi-class anomaly detection, in: Advances in Neural Information Processing Systems (NeurIPS) 35, pp.~4571--4584.
\item Zamir, S.W., Arora, A., Khan, S., Hayat, M., Khan, F.S., Yang, M.-H., 2022. Restormer: Efficient transformer for high-resolution image restoration, in: IEEE/CVF Conference on Computer Vision and Pattern Recognition (CVPR), pp.~5728--5739.
\item Zimmerer, D., Kohl, S.A.A., Petersen, J., Isensee, F., Maier-Hein, K.H., 2018. Context-encoding variational autoencoder for unsupervised anomaly detection. arXiv:1812.05941.
\end{list}
\end{small}

\end{document}